\documentclass{article}

\PassOptionsToPackage{numbers, compress}{natbib}

\usepackage[main, final]{neurips_2026}

\usepackage[utf8]{inputenc} 
\usepackage[T1]{fontenc}    
\usepackage{hyperref}       
\usepackage{url}            
\usepackage{booktabs}       
\usepackage{amsfonts}       
\usepackage{nicefrac}       
\usepackage{microtype}      
\usepackage[table]{xcolor}         
\usepackage{amsmath}
\usepackage{amssymb}
\usepackage{enumitem}
\usepackage{graphicx}
\usepackage{tikz}
\usetikzlibrary{arrows.meta}
\usepackage[normalem]{ulem}
\usepackage{rotating}
\useunder{\uline}{\ul}{}
\usepackage{multirow}
\usepackage{booktabs}
\usepackage{tabularx}
\usepackage{wrapfig}
\usepackage{algorithm}
\usepackage[noend]{algpseudocode}
\algrenewcommand\algorithmicrequire{\textbf{Input:}}
\algrenewcommand\algorithmicensure{\textbf{Output:}}
\algrenewcommand\algorithmiccomment[1]{\hfill$\triangleright$~#1}

\title{SliMOO: Interpretable Multi-Objective Evolutionary Search for LLM Depth Pruning}

\author{
Guanchen Li, Yixing Xu, Xuanwu Yin, Dong Li, Emad Barsoum \\
Advanced Micro Devices, Inc. (AMD) \\
\texttt{\{guanchen.li,yixing.xu,xuanwu.yin,d.li,emad.barsoum\}@amd.com}
}

\begin{document}

\maketitle

\begin{abstract}
    As large language models (LLMs) grow more powerful, their scale makes them costly to serve. Depth pruning offers a practical route to efficiency by directly reducing memory and latency without hardware-specific support. However, we observe that transformer layers exhibit markedly different task-dependent layer contributions, whereas existing depth-pruning methods ignore this heterogeneity and thus tend to optimize for a narrow domain or benchmark rather than preserve broad capability. This suggests that LLM depth pruning should be formulated as a multi-objective optimization problem rather than a single-score ranking problem.
    
    We propose \textbf{SliMOO}, an interpretable multi-objective evolutionary search framework for LLM depth pruning. 
    SliMOO first uses single-layer removal probing to construct a task-layer importance map, which provides an interpretable view of task-shared and task-specific layers and serves as a search prior. 
    It then performs a Pareto-aware evolutionary search for pruning masks, where candidates are generated under the guidance of the task-layer prior, scored by their deviation from the dense model on each task, and retained through NSGA-II environmental selection.
    Across the Llama-3.1 and Qwen3 model families, SliMOO consistently achieves stronger multi-task trade-offs than rule-based, greedy, and scalarized baselines. For example, on Llama-3.1-70B at $25\%$ sparsity, it improves the math-domain average from $24.33\%$ to $34.40\%$ and the code-domain average from $35.03\%$ to $39.14\%$.
    Code will be made available at \url{https://github.com/AMD-AGI/SliMOO}.

\end{abstract}

\section{Introduction}\label{sec:introduction}

Large language models (LLMs) have advanced rapidly in capability~\cite{deepseek,llama3,qwen3}, but their growing scale makes their deployment increasingly costly. This tension has motivated extensive work on model compression, such as pruning, sparsity, and quantization~\cite{tyr,wanda,gptq}. Among these approaches, we focus on depth pruning, which removes entire transformer layers or sublayers from a pretrained model and therefore translates more directly into practical speedups on standard hardware, without requiring specialized sparse kernels or custom accelerators~\cite{sparse-hardware,shortgpt}.

Despite practical advantages, existing layer-pruning methods typically estimate layer importance with a single global score, derived from heuristics~\cite{laco,shortgpt}, calibration loss~\cite{blockpruner}, or scalarized search~\cite{evopress,finercut}. This presumes a universal pruning order across capability domains, whereas we observe markedly different task-dependent contributions across transformer layers, suggesting that different capabilities rely on different depths. Consequently, a pruning strategy that appears successful by a headline metric may still fail to preserve broad capability: on Llama-3.1-70B at $25\%$ sparsity, ShortGPT keeps MMLU, a knowledge reasoning benchmark, nearly unchanged ($77.79$ versus $78.51$ for the dense model), yet sharply degrades GSM8K (math) from $80.82$ to $12.74$ and HumanEval (code) from $51.83$ to $18.29$. This cross-domain divergence indicates that depth pruning for LLMs is inherently a multi-objective optimization problem rather than a single-score ranking problem.

To address this gap, we propose \textbf{SliMOO}\footnote{``Slim'' refers to model slimming through depth pruning, and ``MOO'' stands for multi-objective optimization.}, an interpretable multi-objective evolutionary search framework for LLM depth pruning. The key idea is to move from a single global pruning order to a task-aware multi-objective search over pruning masks. Concretely, SliMOO treats each task as a separate objective, defined by the KL divergence between the dense and pruned models' output distributions on the same task-specific inputs, and searches for masks that preserve strong multi-task trade-offs under a fixed sparsity budget. It also yields an interpretable view of which layers are shared across tasks and which are task-specific. SliMOO realizes this idea with two main contributions:

\begin{itemize}
    \item We introduce \textbf{task-layer importance profiling (TLIP)}, which uses single-layer removal probing to quantify the task-wise contribution of individual layers. TLIP yields an interpretable task-layer importance map that identifies task-shared and task-specific layers, and provides the task-layer prior that guides the subsequent search.

    \item SliMOO performs a \textbf{Pareto-aware evolutionary search} over pruning masks. It generates candidate pruning masks through TLIP prior-guided perturbations and sparsity-preserving crossover, evaluates them on the task-wise objectives, and retains a diverse set of high-quality solutions through NSGA-II environmental selection, which combines non-dominated sorting with diversity preservation. This procedure enables SliMOO to identify pruning masks that achieve stronger multi-task trade-offs than single-score ranking methods.

\end{itemize}

\begin{wrapfigure}{r}{0.62\textwidth}
\vspace{-1em}
\centering
\includegraphics[width=\linewidth]{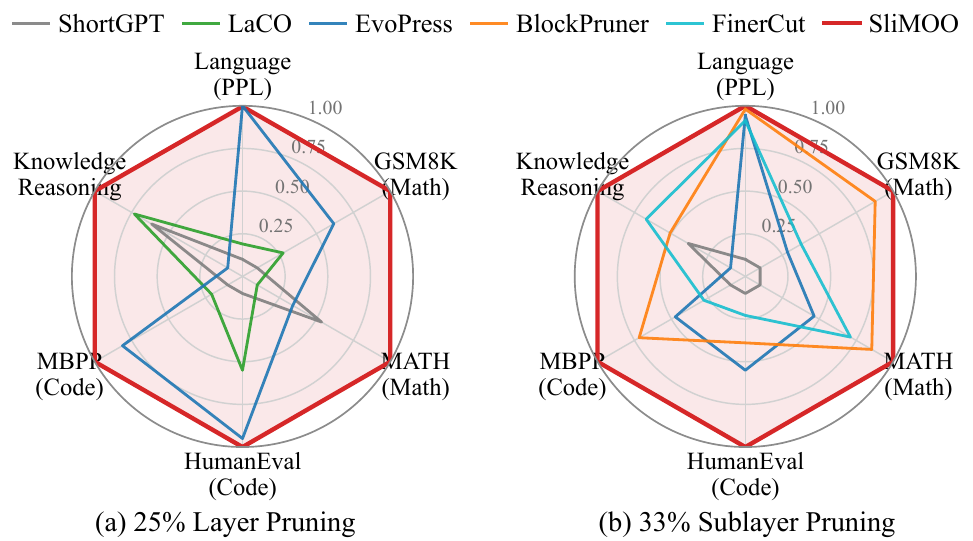}
\caption{Normalized cross-domain trade-offs on Llama-3.1-70B. Scores are normalized within each setting; language modeling perplexity is inverted before normalization.}
\label{fig:llama70b-tradeoff-radar}
\end{wrapfigure}

Together, these components make SliMOO an effective and interpretable framework for LLM depth pruning. Across the Llama-3.1 and Qwen3 model families, SliMOO consistently delivers stronger multi-task trade-offs than rule-based, greedy, and scalarized baselines (\textit{cf.}\ Figure~\ref{fig:llama70b-tradeoff-radar}). For example, on Llama-3.1-70B at $25\%$ sparsity, SliMOO reduces the average language modeling perplexity by $1.46$ relative to ShortGPT and is on par with EvoPress, while improving the math-, code-, and reasoning-domain averages by $21.55/10.07$, $23.69/4.11$, and $0.97/2.27$ relative to ShortGPT and EvoPress, respectively.
\section{Related Work}\label{sec:related-work}

\textbf{Depth pruning for large language models}. Recent work has explored LLM depth pruning to improve deployment efficiency. 
ShortGPT~\cite{shortgpt} ranks transformer layers using an activation cosine similarity score and removes globally redundant layers. 
LaCo~\cite{laco} reduces depth through layer collapse (parameter merging) rather than direct deletion. 
EvoPress~\cite{evopress} applies evolutionary search to identify budget-aware compression configurations, while FinerCut~\cite{finercut} and BlockPruner~\cite{blockpruner} prune finer-grained attention/MLP sublayers using calibrated search or output-change signals. 
These methods demonstrate that substantial depth or structure can be removed, but they typically rely on a single global importance notion or a scalar objective. 
A related line of work skips computation dynamically at inference time through early exit or token-dependent routing, as in CALM~\cite{calm}, LayerSkip~\cite{layerskip}, and Mixture-of-Depths~\cite{mod}. 
Such methods can reduce FLOPs or latency, but because they do not permanently remove layers, they do not reduce checkpoint size or memory footprint, and are often harder to integrate with static serving frameworks such as vLLM~\cite{vllm} and SGLang~\cite{sglang}.

\textbf{Multi-objective optimization}. 
Multi-objective optimization (MOO) has a long history in optimization theory and algorithms, from early Pareto-based genetic algorithms~\cite{moo1,moo2} to comprehensive treatments of multi-objective search~\cite{moo3}. 
NSGA-II~\cite{NSGA2} in particular established non-dominated sorting and diversity-preserving environmental selection as a standard template for approximating Pareto fronts. 
In deep learning, multi-objective search has also been applied to neural architecture search and deployment-aware model design to balance accuracy with resource constraints~\cite{nas-moo}. 
For model compression, MOSP~\cite{mosp} formulates unstructured LLM pruning as a multi-objective problem, using ADMM to identify shared and task-specific support that generate Pareto sparse models.

\textbf{LLM interpretability and task sensitivity}. A growing line of work investigates the functional roles of transformer components at multiple granularities, from layers and heads to neurons and parameters, using probing, ablation, and intervention-based analysis. Probing studies showed that different linguistic phenomena become recoverable at different depths \cite{bert-redis}. Some recent work examines which layers are globally important \cite{layer-importance} and which internal units become specialized for different domains in multi-task settings \cite{coling-best}. Taken together, these studies suggest that contributions to model behavior are unevenly distributed across different components.

\textbf{Positioning of SliMOO.} SliMOO is motivated by a key limitation of existing depth-pruning methods: they do not explicitly account for the task-specific functional roles of layers. Unlike MOSP-style intra-weight support optimization, SliMOO operates on layer/sublayer supports in a discrete depth-pruning space, where structural removals induce inter-layer interactions; its interpretable task-layer profiling guides feasible mask search, making the search both task-aware and trade-off-aware.
\section{Method}\label{sec:method}

\begin{figure}[t]
    \centering
    \includegraphics[width=\linewidth]{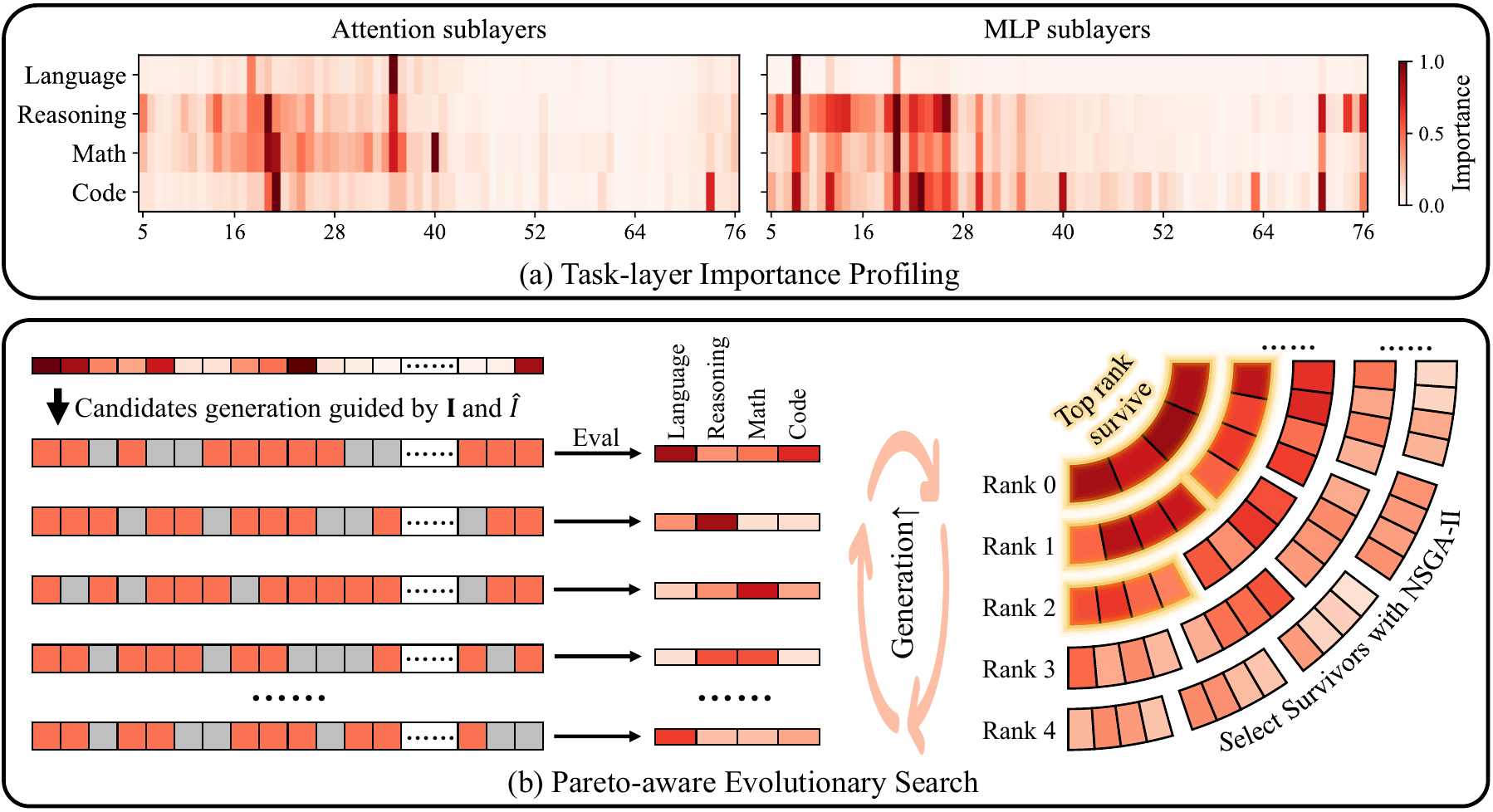}
    \caption{\textbf{Overview of SliMOO.} (a) TLIP profiles task-wise importance of layers/sublayers, yielding a task-layer importance map $\mathbf{I}$. (b) Guided by the task-layer prior, SliMOO generates feasible pruning masks and uses NSGA-II to retain Pareto-competitive trade-offs under a target sparsity.}
    \label{fig:slimoo}
\end{figure}

SliMOO introduces a multi-objective task-aware framework for LLM depth pruning (\textit{cf.}\ Figure~\ref{fig:slimoo}). The method first measures how each removable layer or sublayer affects different capability domains, forming a task-layer importance map. This map is then used as a structured prior that guides candidate generation inside a Pareto-aware evolutionary search under an exact sparsity budget. 
We first define the pruning search space, review the multi-objective optimization preliminaries, and establish our motivation (\textit{cf.}\ \S\ref{subsec:Preliminaries}); 
we then introduce task-layer importance profiling (\textit{cf.}\ \S\ref{subsec:tlip}) and describe the task-wise objective and TLIP-guided Pareto search procedure (\textit{cf.}\ \S\ref{subsec:guided-pareto-search}).

\subsection{Preliminaries}\label{subsec:Preliminaries}

\textbf{Pruning units and search space}. Consider an LLM with $N$ transformer layers. SliMOO supports two depth-pruning granularities: removing an entire transformer layer, or removing individual attention/MLP sublayers, inducing search dimensions of $D=N$ and $D=2N$, respectively. A pruning pattern is represented by a binary mask $\mathbf{x}\in\{0,1\}^{D}$, where $x_i=1$ indicates that unit $i$ is removed. In practice, some units may be excluded from the search a priori: for example, the first and last few layers are commonly more sensitive and are rarely safe pruning targets. We encode the remaining removable units with a feasible set $\mathcal{F}$. Under an exact budget of $K$ removed units, SliMOO searches over:
\begin{equation}\label{eq:search-space}
\mathcal{X}_K=\{\mathbf{x}\in\{0,1\}^{D}\mid \mathrm{supp}(\mathbf{x})\subseteq\mathcal{F},\ \|\mathbf{x}\|_1=K\}.
\end{equation}

\textbf{Multi-objective optimization}.
For a candidate mask $\mathbf{x}$, let $\mathbf{f}(\mathbf{x})=(f_1(\mathbf{x}),\ldots,f_T(\mathbf{x}))$ denote its task-wise pruning losses on $T$ tasks, where all objectives are minimized. Multi-objective optimization seeks masks that are simultaneously good across tasks rather than collapsing heterogeneous capabilities into a single scalar score. A candidate mask $\mathbf{x}$ dominates $\mathbf{y}$ if it is no worse on every objective and strictly better on at least one:
\begin{equation}\label{eq:pareto-dominance}
\mathbf{x}\prec\mathbf{y}
\iff
f_t(\mathbf{x})\le f_t(\mathbf{y})\ \forall t,
\quad
\exists t' : f_{t'}(\mathbf{x})<f_{t'}(\mathbf{y}).
\end{equation}
The corresponding Pareto-optimal set is \( \mathcal P=
\{\mathbf{x}\in\mathcal X_K \mid \not\exists\, \mathbf{y}\in\mathcal X_K:\ \mathbf{y}\prec \mathbf{x}\} \).
The evolutionary search iteratively generates offspring candidates from the current population through variation operators, then selects the next generation from the union of parents and offspring. NSGA-II provides a natural environmental selection mechanism for this process: it partitions candidates into non-dominated fronts, retains lower-rank fronts first, and promotes diversity within the last retained front through the crowding distance:
\begin{equation}\label{eq:crowding-distance}
\mathrm{CD}(\mathbf{x})=\sum_{t=1}^{T}
\frac{f_t(\mathbf{x}_t^{+})-f_t(\mathbf{x}_t^{-})}{f_t^{\max}-f_t^{\min}},
\end{equation}
where $\mathbf{x}_t^{-}$ and $\mathbf{x}_t^{+}$ are the immediate neighbors of candidate $\mathbf{x}$ after sorting the current front by objective $t$, and $f_t^{\max}$/$f_t^{\min}$ are the front-wise maximum/minimum values for normalization. Boundary candidates are assigned infinite distance. This rule favors candidates in sparse regions of the same Pareto front, preserving diverse trade-offs instead of collapsing to masks that optimize only one capability domain.

\textbf{Motivation}.
A single global pruning order would only be appropriate if all tasks depended on similar layers. Figure~\ref{fig:slimoo}(a) shows that this is not the case: for example, the MLP unit at layer index 8 is important across all tasks, whereas the MLP unit at layer index 40 is mainly important for code. SliMOO is therefore designed around two complementary requirements: the search should know which units are risky for each task, and selection should preserve masks that realize different cross-domain trade-offs instead of forcing one scalar preference.

\subsection{Task-layer Importance Profiling}\label{subsec:tlip}

SliMOO begins with \textbf{task-layer importance profiling (TLIP)}, a task-aware probing method for quantifying the contribution of pruning units to different tasks before the Pareto-aware evolutionary search begins. For each removable unit $i$ and task $t$, TLIP removes only that unit and measures the resulting task degradation:
\begin{equation}\label{eq:singleton-effect}
I_{t,i}=f_t(\mathbf e_i)-f_t(\mathbf 0),
\end{equation}
where $\mathbf 0$ denotes the dense model, $\mathbf e_i$ removes only unit $i$ ($f_t$ is defined below). This singleton probe is cheap relative to evaluating all pruning combinations, but it exposes first-order task sensitivity over depth. Repeating it over all units yields a task-layer importance map $\mathbf I\in\mathbb R^{T\times D}$. This map plays a dual role. First, it provides an interpretable view of layer contributions across tasks: each row characterizes one task's sensitivity over depth, while each column indicates whether a unit is broadly shared or selectively important to a particular task. Second, it supplies a family of task-layer priors for Pareto-aware search: the task-specific rows $I_{t,:}$ guide exploration toward individual task trade-offs, while a balanced prior is obtained by conservatively aggregating them:
\begin{equation}\label{eq:task-layer-prior}
\hat I_i=\max_t I_{t,i},
\end{equation}
This max-aggregated prior protects any unit that is critical to at least one task and is therefore used as the risk-averse guide. During search, SliMOO alternates between $\hat I$ and task-specific rows of $\mathbf I$: the former steers pruning away from broadly important units, while the latter injects task-focused variation and helps the population explore different Pareto regions.

\subsection{Pareto-aware Evolutionary Search}\label{subsec:guided-pareto-search}

\textbf{Task-wise pruning objective}. Because our goal is to preserve capability across tasks rather than optimize a single proxy, SliMOO evaluates each pruning mask by how closely the pruned model matches the dense model on each task. This avoids requiring task labels for every calibration token and directly measures distributional distortion caused by pruning. For task $t$, let $c_{n,<j}^{(t)}$ denote the prefix context of the $j$-th token in calibration example $n$, and let $m_{n,j}^{(t)}\in\{0,1\}$ be a token-level loss mask. We define the pruning objective as a masked, entropy-normalized forward KL divergence:
\begin{equation}\label{eq:task-fitness}
f_t(\mathbf{x})
=
\frac{1}{H^{(t)}+\epsilon}
\cdot
\frac{1}{Z_t}
\sum_{n,j}
m_{n,j}^{(t)}
\,
\mathrm{KL}\!\big(
p_0(\cdot \mid c_{n,<j}^{(t)})
\;\|\;
p_{\mathbf{x}}(\cdot \mid c_{n,<j}^{(t)})
\big),
\end{equation}
where $p_0$ and $p_{\mathbf{x}}$ are the dense and pruned next-token distributions, $Z_t=\sum_{n,j}m_{n,j}^{(t)}$, and $H^{(t)}$ is the dense-model entropy averaged over the same active positions. Here, $H^{(t)}$ acts as a task-scale normalizer, so the KL values from high- and low-entropy domains are more comparable. The token mask $m^{(t)}$ enables task-specific scoring, such as response-only masking for math and code evaluation. Collecting these task-wise objectives yields the multi-objective optimization problem:
\begin{equation}\label{eq:moo-pruning-objective}
\min_{\mathbf{x}\in\mathcal X_K}
\mathbf f(\mathbf{x})
=
\big(f_1(\mathbf{x}),\ldots,f_T(\mathbf{x})\big).
\end{equation}
\textbf{TLIP-guided Pareto-aware evolutionary search}. A vanilla NSGA-II search would treat pruning as a black-box vector optimization problem: it generates masks, evaluates them, and selects the next population by Pareto rank and diversity. SliMOO instead injects task-aware structure into candidate generation through TLIP (\textit{cf}. Figure~\ref{fig:slimoo}(b)). TLIP is profiled once before search, and its priors are reused for initialization and mutation. Initialization combines importance-guided masks, which sample $K$ distinct units from $\mathcal F$ without replacement, biased toward low $\hat I$, with random feasible masks for diversity. During evolution, mutation parents are sampled uniformly from the population, crossover uses two distinct uniform parents, and mutations alternate between using $\hat I$ and uniformly sampling one task-specific row from $\{I_{t,:}\}_{t=1}^{T}$ as $\tilde I$. For a parent with pruned set $S\subseteq\mathcal F$, swap mutation samples a kept feasible index $a\in\mathcal F\setminus S$ to prune and a pruned index $r\in S$ to restore:
\begin{equation}\label{eq:swap-kernel}
P_{\mathrm{prune}}(a)\propto \exp(-\tilde I_a/\tau),
\qquad
P_{\mathrm{restore}}(r)\propto \exp(\tilde I_r/\tau),
\end{equation}
where $\tau$ controls exploration versus exploitation: smaller values focus mutation on low-importance units, while larger values flatten sampling. We adapt $\tau$ from population hypervolume~\cite{hv} progress, decaying it after improvement and increasing it after stagnation. In parallel, cardinality-preserving crossover recombines parent pruning sets by inheriting shared removals and filling the remaining budget from their disagreement, so every offspring remains in $\mathcal X_K$. Candidate quality is measured by task-wise objectives on calibration sets rather than held-out evaluation benchmarks. NSGA-II then retains non-dominated and diverse trade-offs, and hypervolume tie-breaking prioritizes candidates with larger Pareto contribution for final selection.
Algorithm~\ref{alg:SliMOO} summarizes the full procedure.

\begin{algorithm}[h!]
    \caption{SliMOO: TLIP-guided Pareto-aware evolutionary search for depth pruning}
    \label{alg:SliMOO}
    \begin{algorithmic}[1]
    \small
    \Statex \textbf{Input:} \parbox[t]{0.8\linewidth}{Dense model $\mathcal{M}$, task calibration sets $\{\mathcal{D}_t\}_{t=1}^{T}$, feasible search space $\mathcal{X}_K$, initial set size $M_0$, population size $P$, generations $G$, offspring budget $B$, crossover ratio $\rho$}
    \Statex \textbf{Output:} \parbox[t]{0.8\linewidth}{Selected pruning mask $\mathbf{x}^{\star}$ and task-layer importance map $\mathbf{I}$}
    \Statex \textbf{Phase 1: Task-layer importance profiling}
    \For{each task $t\in\{1,\dots,T\}$}
        \For{each removable unit $i$}
            \State Compute single-removal effect $I_{t,i}=f_t(\mathbf e_i)-f_t(\mathbf 0)$
        \EndFor
    \EndFor
    \State Form the balanced task-layer prior $\hat I$ with $\hat I_i\gets \max_t I_{t,i}$
    \Statex \textbf{Phase 2: Prior-guided initialization}
    \State Generate $\mathcal{C}_0=\{\mathbf{x}^{(m)}\}_{m=1}^{M_0}$ with importance-guided samples and random feasible masks
    \State Evaluate each $\mathbf{x}\in\mathcal{C}_0$ to obtain its task-wise objective vector $\mathbf{f}(\mathbf{x})$ in Eq.~\eqref{eq:moo-pruning-objective}
    \State Select the initial population $\mathcal{P}_0\gets \textsc{NSGA-II}(\mathcal{C}_0, P)$
    \State Initialize adaptive temperature $\tau$ and offspring sizes $B_{\mathrm{xo}}\gets \lfloor \rho B \rfloor$, $B_{\mathrm{mut}}\gets B-B_{\mathrm{xo}}$
    \Statex \textbf{Phase 3: Pareto evolutionary search}
    \For{$g=1$ to $G$}
        \State Set the parent pool $\mathcal{Q}_g \gets \mathcal{P}_{g-1}$ for uniform parent sampling
        \State Generate $\mathcal{O}_g^{\mathrm{xo}}=\{\mathbf{x}_{g,b}^{\mathrm{xo}}\}_{b=1}^{B_{\mathrm{xo}}}$
        \Statex \hspace{\algorithmicindent}\hspace{\algorithmicindent}by inheriting shared pruned units and filling remaining slots from parent disagreements
        \State Generate $\mathcal{O}_g^{\mathrm{mut}}=\{\mathbf{x}_{g,b}^{\mathrm{mut}}\}_{b=1}^{B_{\mathrm{mut}}}$
        \Statex \hspace{\algorithmicindent}\hspace{\algorithmicindent}by alternating between $\hat I$ and a uniformly sampled task row as $\tilde I$, then applying Eq.~\eqref{eq:swap-kernel}
        \State Form candidate pool $\mathcal{C}_g\gets \mathcal{P}_{g-1}\cup\mathcal{O}_g^{\mathrm{xo}}\cup\mathcal{O}_g^{\mathrm{mut}}$
        \State Evaluate each $\mathbf{x}\in\mathcal{C}_g$ to obtain its task-wise objective vector $\mathbf{f}(\mathbf{x})$ in Eq.~\eqref{eq:moo-pruning-objective} \Comment{reuse parent scores}
        \State Select $\mathcal{P}_g\gets \textsc{NSGA-II}(\mathcal{C}_g, P)$
        \State Adapt $\tau$ from population HV progress
    \EndFor
    \State Select $\mathbf{x}^{\star}\gets \textsc{HV-Best}(\mathcal{P}_G)$
    \State Return $\mathbf{x}^{\star}$, $\mathbf I$
    \end{algorithmic}
    \end{algorithm}
\section{Experiments}\label{sec:experiments}

\subsection{Experimental setup}\label{subsec:experimental-setup}

\textbf{Models and pruning settings}. We evaluate SliMOO on four main large language models: Llama-3.1-70B, Llama-3.1-405B, Qwen3-32B, and Qwen3-235B-A22B. We study two pruning granularities: whole-layer pruning at $25\%$ sparsity and attention/MLP sublayer pruning at $33\%$ sparsity.

\textbf{Calibration}. SliMOO uses four domain-specific sources: \texttt{fineweb\_edu}~\cite{fineweb} for language modeling, \texttt{commonsense\_170k}~\cite{dora} for knowledge reasoning, \texttt{GSM8K} (train split)~\cite{gsm8k} plus \texttt{DeepScaleR-Preview}~\cite{deepscaler} for math, and \texttt{python-codes}~\cite{pythoncodes25k} for code. 
For math and code, we follow a supervised response-only masking scheme and compute the objective only on solution or code-completion tokens.
The multi-domain budget is split evenly across four domains, yielding $65{,}536$ tokens per domain with maximum length $8{,}192$; the math budget is split evenly between \texttt{GSM8K} (easy) and \texttt{DeepScaleR-Preview} (hard).
For the main comparison, calibration-based baselines use the same total budget but the generic source \texttt{fineweb\_edu}; the fair comparison additionally reruns them on SliMOO's mixed multi-domain pool to isolate calibration-source effects.

We compare against representative depth and structured pruning baselines: ShortGPT~\cite{shortgpt}, LaCo~\cite{laco}, EvoPress~\cite{evopress}, BlockPruner~\cite{blockpruner}, FinerCut~\cite{finercut}, and LLM-Pruner~\cite{llm-pruner}. LLM-Pruner, a head/neuron pruning method, is included to broaden the comparison beyond pure depth pruning. LaCo merges LLM layers without calibration, which is evaluated in its original calibration-free setting.

\textbf{Evaluation tasks}.
We evaluate downstream performance on four profiled capability domains:
\begin{enumerate}[leftmargin=2.7em,itemsep=0.2em,topsep=0.0em,parsep=0em,partopsep=0em]
\item \textit{Language modeling:}
WikiText-2~\cite{wikitext2}, PTB~\cite{ptb}, and C4~\cite{c4}
perplexity (lower is better).

\item \textit{Math:}
GSM8K~\cite{gsm8k} (5-shot exact match) and
Hendrycks MATH~\cite{math} (4-shot exact match).

\item \textit{Code:}
HumanEval~\cite{humaneval} (0-shot pass@1) and
MBPP~\cite{mbpp} (3-shot pass@1).

\item \textit{Knowledge reasoning:}
MMLU~\cite{mmlu} (5-shot accuracy), plus
ARC-Challenge, ARC-Easy~\cite{arc}, Social-IQA~\cite{siqa},
HellaSwag~\cite{hellaswag}, PIQA~\cite{piqa}, and
WinoGrande~\cite{winogrande} (0-shot accuracy).
\end{enumerate}

We additionally include five auxiliary benchmarks for out-of-domain evaluation: RULER~\cite{ruler} (NIAH-8K) for long-context retrieval, TruthfulQA~\cite{truthfulqa} (MC2) for truthfulness, CrowS-Pairs~\cite{crows-pairs} for social bias, ToxiGen~\cite{toxigen} for toxicity and implicit hate speech, and MedMCQA~\cite{medmcqa} for medical question answering. These benchmarks are excluded from TLIP and evolutionary search.

\textbf{Implementation details}. 
Our implementation is built with \texttt{PyTorch}~\cite{torch} and uses Hugging Face \texttt{Transformers}~\cite{transformers} and \texttt{Datasets}~\cite{datasets} to manage models and data.
TLIP uses $8{,}192$ profiling tokens per task. For all models, the first two and last two layers are treated as pruning-sensitive and excluded from the search space. SliMOO search runs use $30$ generations, population size $4$, $64$ offspring per generation, $128$ initial candidates, and crossover rate $0.3$. For initialization, $\tau=1.0$: half of the candidates are importance-guided samples with probabilities proportional to $\exp(-\hat I_i/\tau)$, and half are random feasible masks for diversity; when generating offspring, $\tau$ decays by $0.9$ on population HV improvement and grows by $1.5$ after $3$ stagnant generations, clipped to $[0.1,5.0]$.

\subsection{Main results}\label{subsec:main-results}

\begin{table*}[t]
\centering
\caption{Post-pruning performance on the Llama family. We report domain averages: lower is better for language modeling and higher is better for math, code, and knowledge reasoning. \textbf{Bold} indicates the best result, and \underline{underlines} indicate the runner-up.}
\label{tab:llama-main}
\setlength{\tabcolsep}{5pt}
\renewcommand{\arraystretch}{1.03}
\resizebox{\textwidth}{!}{
\begin{tabular}{c|c|l|cccc}
\toprule
Model & Sparsity & Method & Language Avg.\ $\downarrow$ & Math Avg.\ $\uparrow \%$ & Code Avg.\ $\uparrow\%$ & Knowledge Avg.\ $\uparrow\%$ \\
\midrule
\multirow{11}{*}{Llama-3.1-70B}
& \cellcolor{gray!15}$0\%$
& \cellcolor{gray!15}- & \cellcolor{gray!15}5.73 & \cellcolor{gray!15}52.48 & \cellcolor{gray!15}58.32 & \cellcolor{gray!15}72.25 \\ \cmidrule(lr){2-7}
& \multirow{5}{*}{\shortstack{$25\%$\\Layer}}
& ShortGPT & 11.35 & 12.85 & 15.45 & 65.93 \\
& & LaCo & 11.20 & 16.00 & 20.39 & 66.22 \\
& & LLM-Pruner & \textbf{8.74} & \underline{32.44} & 26.78 & \underline{66.78} \\
& & EvoPress & \underline{9.88} & 24.33 & \underline{35.03} & 64.63 \\
& & \cellcolor{yellow!15}SliMOO & \cellcolor{yellow!15}9.89 & \cellcolor{yellow!15}\textbf{34.40} & \cellcolor{yellow!15}\textbf{39.14} & \cellcolor{yellow!15}\textbf{66.90} \\
\cmidrule(lr){2-7}
& \multirow{5}{*}{\shortstack{$33\%$\\Sublayer}}
& ShortGPT & 35.22 & 5.20 & 2.75 & 63.38 \\
& & EvoPress & 12.02 & 10.33 & 15.11 & 62.02 \\
& & BlockPruner & \underline{10.99} & \underline{24.22} & \underline{18.89} & 63.96 \\
& & FinerCut & 12.85 & 13.10 & 7.87 & \underline{64.73} \\
& & \cellcolor{yellow!15}SliMOO & \cellcolor{yellow!15}\textbf{10.45} & \cellcolor{yellow!15}\textbf{27.27} & \cellcolor{yellow!15}\textbf{30.78} & \cellcolor{yellow!15}\textbf{66.29} \\
\midrule
\multirow{10}{*}{Llama-3.1-405B}
& \cellcolor{gray!15}$0\%$
& \cellcolor{gray!15}- & \cellcolor{gray!15}4.68 & \cellcolor{gray!15}60.41 & \cellcolor{gray!15}62.94 & \cellcolor{gray!15}75.01 \\ \cmidrule(lr){2-7}
& \multirow{5}{*}{\shortstack{$25\%$\\Layer}}
& ShortGPT & 8.10 & 28.68 & 48.73 & 70.58 \\
& & LaCo & 8.71 & 40.69 & \underline{49.26} & \underline{70.59} \\
& & EvoPress & \textbf{7.45} & \underline{44.95} & 46.90 & 70.22 \\
& & \cellcolor{yellow!15}SliMOO & \cellcolor{yellow!15}\underline{7.59} & \cellcolor{yellow!15}\textbf{49.83} & \cellcolor{yellow!15}\textbf{51.99} & \cellcolor{yellow!15}\textbf{71.03} \\
\cmidrule(lr){2-7}
& \multirow{5}{*}{\shortstack{$33\%$\\Sublayer}}
& ShortGPT & 22.41 & 1.56 & 0.00 & 33.02 \\
& & EvoPress & 9.26 & 29.32 & 41.25 & 67.10 \\
& & BlockPruner & \underline{7.53} & 42.49 & 40.13 & 69.25 \\
& & FinerCut & 9.97 & \underline{45.51} & \underline{43.81} & \underline{71.13} \\
& & \cellcolor{yellow!15}SliMOO & \cellcolor{yellow!15}\textbf{7.27} & \cellcolor{yellow!15}\textbf{54.57} & \cellcolor{yellow!15}\textbf{59.95} & \cellcolor{yellow!15}\textbf{71.71} \\
\bottomrule
\end{tabular}}
\end{table*}

\textbf{Post-pruning performance}. Table~\ref{tab:llama-main} shows that preserving a single domain is not sufficient for reliable depth pruning. Under the $25\%$ whole-layer sparsity setting, SliMOO consistently improves the strongest multi-task trade-offs across both scales. On Llama-3.1-70B, ShortGPT remains reasonably competitive on the knowledge reasoning-domain average ($65.93\%$), but collapses to $12.85\%$ and $15.45\%$ on the math- and code-domain averages; SliMOO raises these to $34.40\%$ and $39.14\%$ while also achieving the best knowledge reasoning-domain average ($66.90\%$). Relative to the calibration-free LaCo and the evolutionary search baseline EvoPress, SliMOO delivers the largest gains on math and code. 
We also report LLM-Pruner as a structural pruning reference: although it achieves the best language-modeling average on Llama-3.1-70B at $25\%$ sparsity, it remains weaker than SliMOO on math, code, and knowledge reasoning. Its $405$B results are omitted due to memory constraints.

At $33\%$ sparsity, where pruning operates at the attention/MLP sublayer level, the problem becomes harder and SliMOO's advantage becomes clearer. On Llama-3.1-70B, it improves the math/code-domain averages from $10.33\%/15.11\%$ under EvoPress to $27.27\%/30.78\%$. On Llama-3.1-405B, it surpasses FinerCut by $+9.06$ points on math and $+16.14$ points on code, while still achieving the best language modeling average and remaining competitive on knowledge reasoning. The detailed results for both the Llama and Qwen families are reported in Appendix~\ref{detailed}.

\begin{figure}[h!]
    \centering
    \includegraphics[width=0.99\linewidth]{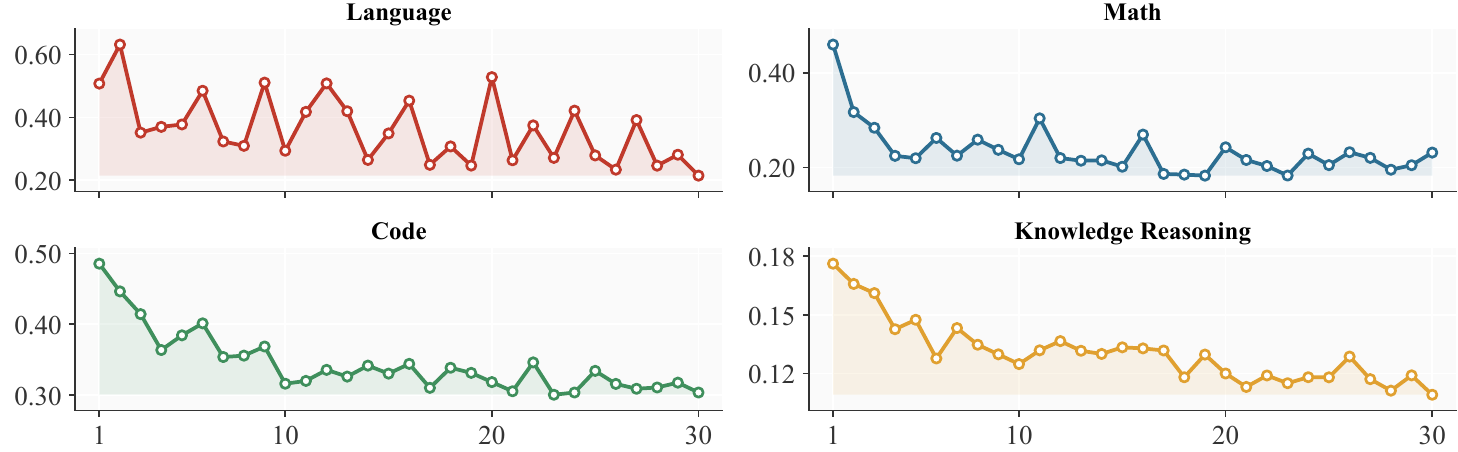}
    \caption{Search dynamics of SliMOO across four capability domains. The horizontal axis denotes the generation index, and the vertical axis reports the corresponding domain-specific search objectives used during optimization; lower values indicate smaller distributional deviation from the dense model.}
    \label{fig:dynamics}
\end{figure}

\textbf{Search dynamics}.
Figure~\ref{fig:dynamics} visualizes the evolution of SliMOO's search objectives across generations. All task-wise objectives drop sharply in the first few generations, showing that TLIP-guided initialization and mutation quickly move the population toward promising pruning regions. The curves then improve more gradually, reflecting the role of Pareto-aware evolution in refining trade-offs rather than simply optimizing the easiest domain. Importantly, no objective shows persistent deterioration or collapse, suggesting that SliMOO maintains balanced progress across language, math, code, and knowledge reasoning throughout search.

\begin{table*}[t]
\centering
\caption{Performance on the four profiled capability domains and five out-of-domain benchmarks for Llama-3.1-70B. \texttt{-mix} denotes calibration using the same mixed-domain
samples and total token budget as SliMOO. All out-of-domain evaluations use the existing pruning mask without additional profiling, search, or fine-tuning. }
\label{tab:unseen-tasks}
\setlength{\tabcolsep}{2.8pt}
\renewcommand{\arraystretch}{1.12}
\footnotesize
\resizebox{\textwidth}{!}{%
\begin{tabular}{@{}c|l|cccc|ccccc@{}}
\toprule
\multirow{3}{*}{Sparsity}
& \multirow{3}{*}{Method}
& \multicolumn{4}{c|}{Profiled capability domains} & \multicolumn{5}{c}{Out-of-domain benchmarks} \\
\cline{3-6} \cline{7-11} \noalign{\vskip 2pt}
&
& Lang. $\downarrow$ & Math $\uparrow$ & Code $\uparrow$ & Know.\ $\uparrow$ & $\vcenter{\hbox{\shortstack{RULER\\NIAH-8K $\uparrow$}}}$ & $\vcenter{\hbox{\shortstack{TruthfulQA\\MC2 $\uparrow$}}}$ & $\vcenter{\hbox{\shortstack{CrowS\\Pairs$^\dagger$}}}$ & ToxiGen $\uparrow$ & MedMCQA $\uparrow$ \\

\midrule

\rowcolor{gray!12}
$0\%$
& Dense
& 5.73
& 52.48
& 58.32
& 72.25
& 100.0
& 51.14
& 68.63
& 46.49
& 69.28 \\

\midrule

\multirow{5}{*}{\shortstack{$25\%$\\Layer}}
& ShortGPT
& 11.35
& 12.85
& 15.45
& 65.93
& 99.8
& 44.80
& 63.80
& 42.55
& 56.30 \\

& ShortGPT\texttt{-mix}
& 11.77
& 13.17
& 19.48
& 65.43
& 99.9
& 45.53
& 64.70
& 43.72
& \underline{60.00} \\

& EvoPress
& \textbf{9.88}
& 24.33
& 35.03
& 64.63
& 100.0
& 42.60
& 65.24
& 43.19
& 54.65 \\

& EvoPress\texttt{-mix}
& 11.35
& \underline{27.89}
& \underline{35.62}
& \underline{66.12}
& 100.0
& \underline{46.27}
& \underline{65.06}
& \underline{44.36}
& 58.04 \\

\rowcolor{yellow!15}
& \textbf{SliMOO}
& \underline{9.89}
& \textbf{34.40}
& \textbf{39.14}
& \textbf{66.90}
& 100.0
& \textbf{48.23}
& \textbf{66.13}
& \textbf{44.68}
& \textbf{62.66} \\

\midrule

\multirow{5}{*}{\shortstack{$33\%$\\Sublayer}}
& EvoPress
& 12.02
& 10.33
& 15.11
& 62.02
& 100.0
& 44.39
& 63.63
& 41.38
& 53.35 \\

& EvoPress\texttt{-mix}
& \underline{10.92}
& \textbf{28.29}
& \underline{24.04}
& 64.46
& 100.0
& \underline{46.14}
& \underline{64.28}
& \underline{42.45}
& \underline{56.06} \\

& FinerCut
& 12.85
& 13.10
& 7.87
& \underline{64.73}
& 100.0
& 43.45
& 62.85
& 38.29
& 52.55 \\

& FinerCut\texttt{-mix}
& 13.41
& 18.18
& 16.32
& 62.65
& 99.9
& 44.92
& 63.51
& 40.32
& 53.26 \\

\rowcolor{yellow!15}
& \textbf{SliMOO}
& \textbf{10.45}
& \underline{27.27}
& \textbf{30.78}
& \textbf{66.29}
& 100.0
& \textbf{46.84}
& \textbf{65.41}
& \textbf{42.66}
& \textbf{56.18} \\

\bottomrule
\end{tabular}%
}

\vspace{0.4em}
\begin{minipage}{\textwidth}
\footnotesize
$^\dagger$ CrowS-Pairs reports stereotypical preference.
Closeness to the dense-model score indicates behavioral
retention.
\end{minipage}

\end{table*}


\subsection{Generalization to Unseen Tasks}
\label{subsec:unseen-tasks}

We evaluate the existing Llama-3.1-70B pruning mask on five out-of-domain benchmarks without additional profiling, search, or fine-tuning, as shown in Table~\ref{tab:unseen-tasks}.

SliMOO achieves the highest TruthfulQA, ToxiGen, and MedMCQA scores among the evaluated pruned models. At $25\%$ layer sparsity, it exceeds the strongest mixed-calibration baseline by $1.96$ and $2.66$ points on TruthfulQA and MedMCQA, respectively. At $33\%$ sublayer sparsity, the corresponding gains are $0.70$ and $0.12$ points, while EvoPress-mix retains a higher in-domain math score ($28.29$ vs.\ $27.27$). RULER-NIAH is nearly saturated across methods, and SliMOO remains closest to the dense model's CrowS-Pairs behavior. These results suggest that balancing heterogeneous in-domain objectives steers pruning toward layers that are broadly redundant across capabilities, helping preserve performance on tasks outside the search objectives.

\subsection{Multilingual Domain Transfer}
\label{broader}

\begin{wraptable}{r}{0.457\textwidth}
\vspace{-1.7em}
\centering
\caption{Multilingual C4 perplexity on Qwen3-32B after $25\%$ layer pruning. Lower is better, and \textbf{bold} marks the best pruned model.}
\label{tab:multilingual-transfer}
\setlength{\tabcolsep}{4pt}
\renewcommand{\arraystretch}{1.08}
\footnotesize
\resizebox{\linewidth}{!}{
\begin{tabular}{l|cccc}
\toprule
Method & EN $\downarrow$ & ES $\downarrow$ & FR $\downarrow$ & DE $\downarrow$ \\
\midrule
\cellcolor{gray!15}Dense & \cellcolor{gray!15}12.19 & \cellcolor{gray!15}9.81 & \cellcolor{gray!15}6.56 & \cellcolor{gray!15}6.94 \\
ShortGPT & 17.75 & 10.56 & 8.75 & 9.31 \\
EvoPress & 16.01 & 10.18 & 8.22 & 8.89 \\
\cellcolor{yellow!15}SliMOO & \cellcolor{yellow!15}\textbf{14.69} & \cellcolor{yellow!15}\textbf{9.81} & \cellcolor{yellow!15}\textbf{7.62} & \cellcolor{yellow!15}\textbf{8.62} \\
\bottomrule
\end{tabular}}
\vspace{-1.0em}
\end{wraptable}

To test whether SliMOO remains useful beyond the capability domains used in the main experiments, we construct a multilingual pruning scenario on Qwen3-32B, a choice with native multilingual support. The calibration budget matches the main setting but is redistributed across \texttt{FineWeb-Edu} English and \texttt{FineWeb-2} Spanish, French, and German subsets~\cite{fineweb2}. Evaluation uses per-token perplexity on C4 multilingual validation splits for English, Spanish, French, and German, with $64$ windows of $4096$ tokens per language. As shown in Table~\ref{tab:multilingual-transfer}, all pruning methods increase perplexity, but SliMOO consistently yields the lowest perplexity among pruned models across the four languages. This suggests that SliMOO is not tied to a fixed domain taxonomy: language/math/code/knowledge and multilingual partitions are examples of how its task-aware multi-objective formulation can be instantiated.

\subsection{Ablation Study}

\textbf{Fair calibration comparison}. Table~\ref{tab:fair-calibration} reports a controlled comparison where all calibration-based baselines are rerun with the same mixed multi-domain calibration data and total token budget as SliMOO. The same conclusion remains: SliMOO's gains cannot be attributed to calibration allocation alone. At $25\%$ layer sparsity, SliMOO improves the math/code averages from $27.89\%/35.62\%$ under EvoPress to $34.40\%/39.14\%$ and achieves the best knowledge average ($66.90\%$). At $33\%$ sublayer sparsity, it attains the best language/code/knowledge averages ($10.45/30.78\%/66.29\%$) while remaining competitive on math ($27.27\%$ versus $28.29\%$ under EvoPress), yielding the strongest overall trade-off. This further supports that SliMOO's advantage comes from its task-wise objective and Pareto-guided search rather than from task-aware calibration allocation alone.

\begin{table*}[t]
\centering
\begin{minipage}[t]{0.53\textwidth}
\vspace{0pt}
\refstepcounter{table}\label{tab:fair-calibration}
\footnotesize
\raggedright
\textbf{Table~\thetable:} Fair calibration comparison on Llama-3.1-70B. All baselines are rerun with the same mixed multi-domain calibration data and total token budget as SliMOO.
\par\centering
\setlength{\tabcolsep}{4pt}
\renewcommand{\arraystretch}{1.03}
\resizebox{\linewidth}{!}{
\begin{tabular}{c|l|cccc}
\toprule
Sp. & Method & Lang. $\downarrow$ & Math $\uparrow$ & Code $\uparrow$ & Know. $\uparrow$ \\
\midrule
\multirow{4}{*}{\shortstack{$25\%$\\Layer}}
& ShortGPT & 11.77 & 13.17 & 19.48 & 65.43 \\
& LLM-Pruner & \textbf{9.42} & \underline{32.44} & 32.30 & \underline{66.15} \\
& EvoPress & 11.35 & 27.89 & \underline{35.62} & 66.12 \\
& \cellcolor{yellow!15}SliMOO & \cellcolor{yellow!15}\underline{9.89} & \cellcolor{yellow!15}\textbf{34.40} & \cellcolor{yellow!15}\textbf{39.14} & \cellcolor{yellow!15}\textbf{66.90} \\
\midrule
\multirow{5}{*}{\shortstack{$33\%$\\Sublayer}}
& ShortGPT & 20.53 & 8.71 & 3.85 & 64.18 \\
& EvoPress & \underline{10.92} & \textbf{28.29} & \underline{24.04} & 64.46 \\
& BlockPruner & 11.38 & 24.16 & 21.43 & \underline{65.40} \\
& FinerCut & 13.41 & 18.18 & 16.32 & 62.65 \\
& \cellcolor{yellow!15}SliMOO & \cellcolor{yellow!15}\textbf{10.45} & \cellcolor{yellow!15}\underline{27.27} & \cellcolor{yellow!15}\textbf{30.78} & \cellcolor{yellow!15}\textbf{66.29} \\
\bottomrule
\end{tabular}}
\end{minipage}\hfill
\begin{minipage}[t]{0.44\textwidth}
\vspace{0pt}
\refstepcounter{table}\label{tab:calibration-ablation}
\footnotesize
\raggedright
\textbf{Table~\thetable:} Calibration budget and allocation ablation for SliMOO on Llama-3.1-70B at $25\%$ layer sparsity. In the dominant (dom.) setting, one domain receives half of the total budget and the rest is evenly split.
\par\centering
\setlength{\tabcolsep}{3.5pt}
\renewcommand{\arraystretch}{1.03}
\resizebox{\linewidth}{!}{
\begin{tabular}{l|cccc}
\toprule
Calib. & Lang. $\downarrow$ & Math $\uparrow$ & Code $\uparrow$ & Know. $\uparrow$ \\
\midrule
\cellcolor{yellow!15}Default ($64$K) & \cellcolor{yellow!15}\underline{9.89} & \cellcolor{yellow!15}\underline{34.40} & \cellcolor{yellow!15}39.14 & \cellcolor{yellow!15}\underline{66.90} \\
$32$K/domain & \underline{9.89} & \underline{34.40} & 39.14 & \underline{66.90} \\
$16$K/domain & 10.13 & 30.25 & \textbf{40.19} & 66.82 \\
$8$K/domain & 10.39 & 29.04 & 37.74 & 66.40 \\
\midrule
Lang.-dom. & \textbf{9.81} & 29.27 & 34.44 & 65.47 \\
Math-dom. & \underline{9.89} & \textbf{35.16} & 28.46 & 65.85 \\
Code-dom. & 10.15 & 18.29 & \underline{39.26} & 66.15 \\
Know.-dom. & 9.97 & 20.04 & 25.07 & \textbf{67.69} \\
\bottomrule
\end{tabular}}
\end{minipage}
\end{table*}

\begin{table*}[t]
\centering
\begin{minipage}[t]{0.53\textwidth}
\vspace{0pt}
\refstepcounter{table}\label{tab:moo-ablation}
\footnotesize
\raggedright
\textbf{Table~\thetable:} Ablation of search objectives and Pareto selection on Llama-3.1-70B at $25\%$ layer sparsity.
\par\centering
\setlength{\tabcolsep}{3.5pt}
\renewcommand{\arraystretch}{1.042}
\resizebox{\linewidth}{!}{
\begin{tabular}{l|cccc}
\toprule
Search variant & Lang. $\downarrow$ & Math $\uparrow$ & Code $\uparrow$ & Know. $\uparrow$ \\
\midrule
Language Evolution & \textbf{9.88} & 24.33 & 35.03 & 64.63 \\
Math Evolution & 10.75 & \textbf{36.11} & 34.89 & 66.44 \\
Code Evolution & \underline{9.89} & 30.93 & \underline{38.59} & 65.66 \\
Knowledge Evolution & 11.35 & 25.46 & 27.20 & \textbf{67.45} \\
\midrule
Mixed Evolution & 11.35 & 27.89 & 35.62 & 66.12 \\
Mean Evolution & 9.97 & 29.14 & 33.10 & 66.62 \\
Random Pareto Evolution & 9.96 & 29.06 & 33.11 & 66.27 \\
HV Evolution & 9.91 & 30.77 & 34.73 & 66.58 \\
NSGA-II & \underline{9.89} & 32.67 & 36.02 & 66.71 \\
\cellcolor{yellow!15}SliMOO & \cellcolor{yellow!15}\underline{9.89} & \cellcolor{yellow!15}\underline{34.40} & \cellcolor{yellow!15}\textbf{39.14} & \cellcolor{yellow!15}\underline{66.90} \\
\bottomrule
\end{tabular}}
\end{minipage}\hfill
\begin{minipage}[t]{0.45\textwidth}
\vspace{0pt}
\refstepcounter{table}\label{tab:hyperparameter-ablation}
\footnotesize
\raggedright
\textbf{Table~\thetable:} Ablation of key SliMOO hyperparameters on Llama-3.1-70B at $25\%$ layer sparsity.
\par\centering
\setlength{\tabcolsep}{4pt}
\renewcommand{\arraystretch}{1.04}
\resizebox{\linewidth}{!}{
\begin{tabular}{l|cccc}
\toprule
Configuration & Lang. $\downarrow$ & Math $\uparrow$ & Code $\uparrow$ & Know. $\uparrow$ \\
\midrule
\cellcolor{yellow!15}SliMOO & \cellcolor{yellow!15}\underline{9.89} & \cellcolor{yellow!15}34.40 & \cellcolor{yellow!15}39.14 & \cellcolor{yellow!15}\underline{66.90} \\
\midrule
20 Generations & 12.10 & 29.49 & 35.77 & 66.36 \\
40 Generations & 10.03 & \underline{35.44} & 37.92 & 66.57 \\ \midrule
32 Offspring & 10.25 & 32.56 & 35.79 & 66.18 \\
128 Offspring & 10.07 & \textbf{35.64} & \underline{40.21} & 65.77 \\
\midrule
Crossover Rate 0.1 & 10.18 & 33.20 & 37.71 & \textbf{67.03} \\
Crossover Rate 0.5 & \underline{9.89} & 34.40 & 39.14 & \underline{66.90} \\ \midrule
Population Size 6 & \textbf{9.78} & 33.08 & \textbf{40.61} & 66.79 \\ \midrule
Fixed temp. $\tau=1$ & 10.44 & 32.61 & 37.27 & 66.37 \\
\bottomrule
\end{tabular}}
\end{minipage}
\end{table*}

\textbf{Calibration budget and allocation}. Table~\ref{tab:calibration-ablation} shows that SliMOO is robust to calibration size. Halving the budget from $64$K to $32$K tokens per domain recovers the same pruning mask and identical post-pruning scores ($9.89/34.40\%/39.14\%/66.90\%$ on language/math/code/knowledge), indicating that the task-layer prior is already stable at moderate calibration sizes. Further reducing the budget to $16$K or $8$K introduces some degradation, but the resulting models remain competitive rather than collapsing, suggesting that SliMOO does not rely on a large calibration set. The allocation study shows a complementary effect: domain-dominant calibration can steer the selected mask toward that domain, with math-dominant calibration giving the best math average ($35.16\%$) and knowledge-dominant calibration giving the best knowledge average ($67.69\%$) while trading off other domains. Adjusting domain proportions is therefore a practical way to reflect user preferences, while equal allocation remains the default for balanced multi-domain pruning.

\textbf{Search objective and Pareto selection}. Table~\ref{tab:moo-ablation} first shows that single-objective evolutionary search improves its target domain but fails to preserve the full capability profile. The multi-objective variants are more balanced but still incomplete: \textit{Mixed Evolution} applies evolutionary search on mixed calibration data, \textit{Mean Evolution} computes each task objective separately and averages them into one scalar, \textit{Random Pareto Evolution} first filters the non-dominated candidates and then samples from them uniformly, and \textit{HV Evolution} keeps candidates with larger hypervolume contribution. 
These alternatives improve over single-domain search but remain weaker than NSGA-II, which explicitly retains non-dominated and diverse candidates. SliMOO further adds TLIP profiling to guide candidate generation toward task-relevant regions, yielding the strongest overall trade-off across domains.

\textbf{Ablation on SliMOO search hyperparameters}. Table~\ref{tab:hyperparameter-ablation} shows that SliMOO is largely robust: changing generations, offspring, crossover, or population size does not cause collapse and still yields competitive trade-offs. Reducing the search to $20$ generations weakens all domains, while $40$ generations improves math but does not dominate the default balance. Increasing the offspring budget to $128$ further improves math and code but hurts knowledge reasoning, indicating a different Pareto point rather than a uniformly better solution. Crossover is stable: rates $0.3$ and $0.5$ recover the same mask, while a smaller rate slightly favors knowledge. A larger population improves language and code but sacrifices math. Finally, fixed $\tau$ degrades the trade-off, suggesting that dynamic exploration helps maintain a balanced Pareto search.

\subsection{Efficiency Analysis}

\begin{wraptable}{r}{0.48\textwidth}
\vspace{-1.7em}
\centering
\caption{Inference efficiency of pruned Llama-3.1-70B with vLLM at batch size 8.}
\label{tab:post-pruning-efficiency}
\setlength{\tabcolsep}{3.5pt}
\renewcommand{\arraystretch}{1.03}
\footnotesize
\resizebox{\linewidth}{!}{
\begin{tabular}{c|c|c|cc}
\toprule
Sparsity & Arch & IO Tok. & TTFT (s) $\downarrow$ & Decode TPS $\uparrow$ \\
\midrule
\multirow{2}{*}{$0\%$} & 80 Attn & 4096 & 3.39 & 482.9 \\
 & 80 FFN & 8192 & 4.47 & 413.6 \\
\midrule
\multirow{2}{*}{$25\%$} & 60 Attn & 4096 & 2.54 & 623.9 \\
 & 60 FFN & 8192 & 3.35 & 539.5 \\
\midrule
\multirow{2}{*}{$33\%$} & 46 Attn & 4096 & 2.37 & 672.2 \\
 & 64 FFN & 8192 & 3.07 & 593.0 \\
\bottomrule
\end{tabular}}
\vspace{-1.0em}
\end{wraptable}

\textbf{Efficiency of SliMOO framework}.
All experiments are conducted on AMD Instinct MI300X GPUs. On Llama-3.1-70B with 2 GPUs, TLIP profiling takes 12.0 minutes and SliMOO search takes 20.8 minutes per generation, close to EvoPress (18.8 minutes $\times$ 30 generations) while producing stronger masks. ShortGPT is fast because it only requires several forward passes, but its post-pruning quality is substantially weaker; FinerCut is much slower, taking 39.2 hours to finish. For Llama-3.1-405B, SliMOO runs on 8 GPUs with a 933\,GB memory budget; profiling takes 46.4 minutes and search takes 52.1 minutes per generation. Thus, SliMOO's one-time offline cost remains acceptable given its best post-pruning trade-offs.

\textbf{Efficiency of post-pruning LLMs}.
Table~\ref{tab:post-pruning-efficiency} reports time to first token (TTFT) and decode throughput (Decode TPS), showing that SliMOO converts structural pruning into practical serving gains. SliMOO prunes the full-depth Llama-3.1-70B model into layer- or sublayer-sparse variants, enabling more efficient serving with standard vLLM. At 4096 IO tokens (equal input and output tokens), $25\%$ layer pruning gives a $1.33\times$ TTFT speedup and $1.29\times$ Decode TPS improvement, while $33\%$ sublayer pruning increases these gains to $1.43\times$ and $1.39\times$. Similar gains hold at 8192 IO tokens, confirming faster inference without specialized sparse kernels.

\section{Limitations}\label{sec:limitations}

SliMOO achieves strong and interpretable cross-domain trade-offs, yet it still has limitations. 
Concretely, the current instantiation of SliMOO uses a predefined set of capability domains and calibration corpora, which locally aligns it with deployment settings whose target domains can be reasonably specified, while broader open-ended, in-the-wild task mixtures merit further study.
SliMOO also incurs a one-time cost for TLIP profiling and evolutionary search, so reducing this overhead would improve practicality. 
For sublayer pruning, our comparisons follow the same structural-unit budget across methods, while attention and MLP units can differ in parameter and FLOP counts; nevertheless, vLLM measurements show that higher-sparsity sublayer pruning can yield noticeably stronger serving speedups than low-sparsity layer pruning even when their parameter counts are relatively close.

\section{Conclusion}

This paper presents \textbf{SliMOO}, an interpretable multi-objective framework for LLM depth pruning. By profiling task-wise layer importance and implementing Pareto-aware evolutionary search, SliMOO avoids the limitations of a single global pruning order and instead identifies pruning masks with stronger cross-domain trade-offs under fixed sparsity budgets. Across the Llama-3.1 and Qwen3 families, SliMOO significantly improves post-pruning performance in the domains of language modeling, math, code, and knowledge reasoning. These results suggest that task-aware multi-objective optimization is a promising direction for reliable and interpretable LLM depth pruning.

\begin{ack}
This work received no external funding. The authors declare no additional competing interests beyond the affiliations disclosed in the paper.
\end{ack}

{\small
\bibliographystyle{IEEEtran}
\bibliography{reference}
}

\newpage
\appendix
\section{Appendix}

\subsection{SliMOO Search Details and Derivation}
\label{slimoo-formula}

The main text defines the feasible mask set $\mathcal{X}_K$, the task-wise KL objectives $\mathbf f(\mathbf x)$, and the TLIP-guided search procedure. This appendix gives the derivation behind these design choices. The central difficulty is that depth pruning is a fixed-budget combinatorial problem with task-dependent and non-additive effects: removing one layer can be harmless for language modeling but harmful for math, and the effect of removing several layers is not exactly the sum of their isolated effects. Let $t$ index a calibration task, $i\in\{1,\ldots,D\}$ index a removable unit, and $x_i\in\{0,1\}$ indicate whether unit $i$ is pruned. If layer interactions were ignored, the task loss of a mask could be approximated by the singleton expansion:
\begin{equation}
\tilde f_t(\mathbf{x}) = \sum_{i=1}^{D} x_i\, I_{t,i},
\end{equation}
where $\tilde f_t(\mathbf{x})$ is the additive approximation to the true task loss $f_t(\mathbf{x})$, and $I_{t,i}=f_t(\mathbf e_i)-f_t(\mathbf 0)$ is the TLIP score obtained by pruning only unit $i$ with the singleton mask $\mathbf e_i$. This additive view explains why TLIP is informative: it estimates the first-order discrete sensitivity of each task to each removable unit. It also explains why TLIP alone is insufficient as a pruning rule. Since transformer layers interact, the best $K$ units to remove cannot be determined reliably by sorting singleton scores alone. SliMOO therefore uses TLIP as a search prior rather than as the final objective: the prior narrows exploration toward plausible masks, while the actual candidate quality is still measured by the full vector objective $\mathbf f(\mathbf x)$.

\begin{sidewaystable}
\centering
\caption{Detailed post-pruning results on the Llama and Qwen families. We report benchmark-level scores together with domain averages: lower is better for language modeling perplexity and higher is better elsewhere. \textbf{Bold} indicates the best result, and \underline{underlines} indicate the runner-up.}
\label{tab:detailed-post-pruning}
\renewcommand{\arraystretch}{1.08}     
\resizebox{0.98\textwidth}{!}{
\begin{tabular}{c|c|l|cccc|ccc|ccc|cccccccc}
\toprule
\multirow{2}{*}{\textbf{Model}} & \multirow{2}{*}{\textbf{Sparsity}} & \multicolumn{1}{c}{\multirow{2}{*}{\textbf{Method}}} & \multicolumn{4}{c}{\textbf{language modeling}} & \multicolumn{3}{c}{\textbf{math}} & \multicolumn{3}{c}{\textbf{code}} & \multicolumn{8}{c}{\textbf{Knowledge Reasoning}} \\ \cmidrule{4-21} 
 &  & \multicolumn{1}{c}{} & \textbf{Wikitext} & \textbf{PTB} & \textbf{C4} & \cellcolor{orange!15}\textbf{AVG} & \textbf{GSM8K} & \textbf{MATH} & \cellcolor{orange!15}\textbf{AVG} & \textbf{HumanEval} & \textbf{MBPP} & \cellcolor{orange!15}\textbf{AVG} & \textbf{MMLU} & \textbf{ARC-C} & \textbf{ARC-E} & \textbf{SIQA} & \textbf{HellaSwag} & \textbf{PIQA} & \textbf{WinoGrande} & \cellcolor{orange!15}\textbf{AVG} \\ \midrule
\multirow{11}{*}{\textbf{Llama-3.1-70B}} & \cellcolor{gray!15}\textbf{0\%} & \cellcolor{gray!15}\textbf{N/A} & \cellcolor{gray!15}2.63 & \cellcolor{gray!15}7.72 & \cellcolor{gray!15}6.84 & \cellcolor{orange!15}5.73 & \cellcolor{gray!15}80.82 & \cellcolor{gray!15}24.14 & \cellcolor{orange!15}52.48 & \cellcolor{gray!15}51.83 & \cellcolor{gray!15}64.80 & \cellcolor{orange!15}58.32 & \cellcolor{gray!15}78.51 & \cellcolor{gray!15}60.75 & \cellcolor{gray!15}87.12 & \cellcolor{gray!15}50.46 & \cellcolor{gray!15}66.43 & \cellcolor{gray!15}82.86 & \cellcolor{gray!15}79.64 & \cellcolor{orange!15}72.25 \\ \cmidrule{2-21} 
 & \multirow{5}{*}{\textbf{\shortstack{25\%\\Layer}}} & \textbf{ShortGPT} & 9.00 & 11.47 & 13.57 & \cellcolor{orange!15}11.35 & 12.74 & 12.96 & \cellcolor{orange!15}12.85 & 18.29 & 12.60 & \cellcolor{orange!15}15.45 & 77.79 & 48.46 & 76.56 & 47.95 & 57.49 & 76.28 & 76.95 & \cellcolor{orange!15}65.93 \\
 &  & \textbf{LaCo} & 8.90 & 11.60 & 13.10 & \cellcolor{orange!15}11.20 & 20.77 & 11.22 & \cellcolor{orange!15}16.00 & 23.78 & 17.00 & \cellcolor{orange!15}20.39 & 78.21 & 48.12 & 76.34 & 47.59 & 58.11 & 76.82 & 78.37 & \cellcolor{orange!15}66.22 \\
 &  & \textbf{LLM-Pruner} & 6.12 & 9.71 & 10.38 & \cellcolor{orange!15}\textbf{8.74} & 50.64 & 14.24 & \cellcolor{orange!15}32.44 & 21.95 & 31.60 & \cellcolor{orange!15}26.78 & 66.72 & 54.10 & 82.11 & 49.23 & 59.85 & 80.58 & 74.90 & \cellcolor{orange!15}66.78 \\
 &  & \textbf{EvoPress} & 7.16 & 10.82 & 11.67 & \cellcolor{orange!15}9.88 & 36.47 & 12.18 & \cellcolor{orange!15}24.33 & 28.66 & 41.40 & \cellcolor{orange!15}35.03 & 66.69 & 50.34 & 77.40 & 46.01 & 57.08 & 78.78 & 76.09 & \cellcolor{orange!15}64.63 \\
 &  & \cellcolor{yellow!15}\textbf{SliMOO} & \cellcolor{yellow!15}7.21 & \cellcolor{yellow!15}10.65 & \cellcolor{yellow!15}11.81 & \cellcolor{orange!15}{\ul 9.89} & \cellcolor{yellow!15}53.98 & \cellcolor{yellow!15}14.82 & \cellcolor{orange!15}\textbf{34.40} & \cellcolor{yellow!15}29.27 & \cellcolor{yellow!15}49.00 & \cellcolor{orange!15}\textbf{39.14} & \cellcolor{yellow!15}74.12 & \cellcolor{yellow!15}52.99 & \cellcolor{yellow!15}79.80 & \cellcolor{yellow!15}46.98 & \cellcolor{yellow!15}58.60 & \cellcolor{yellow!15}78.76 & \cellcolor{yellow!15}77.03 & \cellcolor{orange!15}\textbf{66.90} \\ \cmidrule{2-21} 
 & \multirow{5}{*}{\textbf{\shortstack{33\%\\Sublayer}}} & \textbf{ShortGPT} & 38.28 & 33.54 & 33.83 & \cellcolor{orange!15}35.22 & 4.25 & 6.14 & \cellcolor{orange!15}5.20 & 5.49 & 0.00 & \cellcolor{orange!15}2.75 & 73.00 & 45.82 & 72.73 & 47.49 & 54.67 & 73.83 & 76.09 & \cellcolor{orange!15}63.38 \\
 &  & \textbf{EvoPress} & 9.31 & 12.94 & 13.81 & \cellcolor{orange!15}{\ul 12.02} & 12.13 & 8.52 & \cellcolor{orange!15}10.33 & 14.02 & 16.20 & \cellcolor{orange!15}15.11 & 66.56 & 44.62 & 74.54 & 45.24 & 53.31 & 75.46 & 74.43 & \cellcolor{orange!15}62.02 \\
 &  & \textbf{BlockPruner} & 7.89 & 11.66 & 13.42 & \cellcolor{orange!15}10.99 & 37.38 & 11.06 & \cellcolor{orange!15}24.22 & 10.98 & 26.80 & \cellcolor{orange!15}18.89 & 67.77 & 50.77 & 79.08 & 46.52 & 58.06 & 78.40 & 67.09 & \cellcolor{orange!15}63.96 \\
 &  & \textbf{FinerCut} & 9.33 & 15.15 & 14.06 & \cellcolor{orange!15}12.85 & 16.07 & 10.12 & \cellcolor{orange!15}13.10 & 7.93 & 7.80 & \cellcolor{orange!15}7.87 & 67.29 & 50.26 & 76.26 & 46.62 & 58.22 & 77.69 & 76.79 & \cellcolor{orange!15}64.73 \\
 &  & \cellcolor{yellow!15}\textbf{SliMOO} & \cellcolor{yellow!15}7.34 & \cellcolor{yellow!15}11.81 & \cellcolor{yellow!15}12.19 & \cellcolor{orange!15}\textbf{10.45} & \cellcolor{yellow!15}42.53 & \cellcolor{yellow!15}12.00 & \cellcolor{orange!15}\textbf{27.27} & \cellcolor{yellow!15}22.56 & \cellcolor{yellow!15}39.00 & \cellcolor{orange!15}\textbf{30.78} & \cellcolor{yellow!15}71.38 & \cellcolor{yellow!15}53.41 & \cellcolor{yellow!15}79.50 & \cellcolor{yellow!15}46.47 & \cellcolor{yellow!15}57.77 & \cellcolor{yellow!15}78.67 & \cellcolor{yellow!15}76.80 & \cellcolor{orange!15}\textbf{66.29} \\ \midrule
\multirow{10}{*}{\textbf{Llama-3.1-405B}} & \cellcolor{gray!15}\textbf{0\%} & \cellcolor{gray!15}\textbf{N/A} & \cellcolor{gray!15}1.36 & \cellcolor{gray!15}6.94 & \cellcolor{gray!15}5.75 & \cellcolor{orange!15}4.68 & \cellcolor{gray!15}88.40 & \cellcolor{gray!15}32.42 & \cellcolor{orange!15}60.41 & \cellcolor{gray!15}54.88 & \cellcolor{gray!15}71.00 & \cellcolor{orange!15}62.94 & \cellcolor{gray!15}85.02 & \cellcolor{gray!15}63.05 & \cellcolor{gray!15}86.78 & \cellcolor{gray!15}53.17 & \cellcolor{gray!15}70.33 & \cellcolor{gray!15}84.00 & \cellcolor{gray!15}82.71 & \cellcolor{orange!15}75.01 \\ \cmidrule{2-21} 
 & \multirow{4}{*}{\textbf{\shortstack{25\%\\Layer}}} & \textbf{ShortGPT} & 5.48 & 8.85 & 9.98 & \cellcolor{orange!15}8.10 & 38.67 & 18.68 & \cellcolor{orange!15}28.68 & 40.85 & 56.60 & \cellcolor{orange!15}48.73 & 83.81 & 54.86 & 81.31 & 49.07 & 63.64 & 80.14 & 81.22 & \cellcolor{orange!15}70.58 \\
 &  & \textbf{LaCo} & 5.86 & 9.27 & 10.99 & \cellcolor{orange!15}8.71 & 58.07 & 23.30 & \cellcolor{orange!15}40.69 & 45.12 & 53.40 & \cellcolor{orange!15}49.26 & 83.66 & 55.23 & 82.11 & 49.52 & 63.17 & 80.74 & 79.72 & \cellcolor{orange!15}70.59 \\
 &  & \textbf{EvoPress} & 4.22 & 8.50 & 9.62 & \cellcolor{orange!15}\textbf{7.45} & 69.45 & 20.44 & \cellcolor{orange!15}44.95 & 37.80 & 56.00 & \cellcolor{orange!15}46.90 & 77.63 & 57.76 & 82.83 & 50.15 & 63.26 & 81.12 & 78.77 & \cellcolor{orange!15}70.22 \\
 &  & \cellcolor{yellow!15}\textbf{SliMOO} & \cellcolor{yellow!15}4.22 & \cellcolor{yellow!15}8.75 & \cellcolor{yellow!15}9.81 & \cellcolor{orange!15}{\ul 7.59} & \cellcolor{yellow!15}76.95 & \cellcolor{yellow!15}22.70 & \cellcolor{orange!15}\textbf{49.83} & \cellcolor{yellow!15}48.78 & \cellcolor{yellow!15}55.20 & \cellcolor{orange!15}\textbf{51.99} & \cellcolor{yellow!15}81.69 & \cellcolor{yellow!15}57.00 & \cellcolor{yellow!15}83.59 & \cellcolor{yellow!15}49.28 & \cellcolor{yellow!15}64.12 & \cellcolor{yellow!15}81.77 & \cellcolor{yellow!15}79.79 & \cellcolor{orange!15}\textbf{71.03} \\ \cmidrule{2-21} 
 & \multirow{5}{*}{\textbf{\shortstack{33\%\\Sublayer}}} & \textbf{ShortGPT} & 31.27 & 16.45 & 19.50 & \cellcolor{orange!15}22.41 & 3.11 & 0.00 & \cellcolor{orange!15}1.56 & 0.00 & 0.00 & \cellcolor{orange!15}0.00 & 22.95 & 19.97 & 26.56 & 34.08 & 25.88 & 51.47 & 50.20 & \cellcolor{orange!15}33.02 \\
 &  & \textbf{EvoPress} & 5.97 & 9.81 & 12.00 & \cellcolor{orange!15}9.26 & 42.61 & 16.02 & \cellcolor{orange!15}29.32 & 30.49 & 52.00 & \cellcolor{orange!15}41.25 & 75.82 & 50.51 & 79.50 & 48.26 & 60.29 & 79.49 & 75.85 & \cellcolor{orange!15}67.10 \\
 &  & \textbf{BlockPruner} & 5.00 & 8.42 & 9.18 & \cellcolor{orange!15}7.53 & 66.34 & 18.64 & \cellcolor{orange!15}42.49 & 28.66 & 51.60 & \cellcolor{orange!15}40.13 & 78.66 & 54.95 & 81.48 & 54.91 & 61.65 & 79.65 & 73.48 & \cellcolor{orange!15}69.25 \\
 &  & \textbf{FinerCut} & 6.04 & 10.72 & 13.15 & \cellcolor{orange!15}9.97 & 69.45 & 21.56 & \cellcolor{orange!15}45.51 & 39.02 & 48.60 & \cellcolor{orange!15}43.81 & 82.54 & 56.91 & 82.70 & 50.77 & 63.41 & 80.58 & 80.98 & \cellcolor{orange!15}71.13 \\
 &  & \cellcolor{yellow!15}\textbf{SliMOO} & \cellcolor{yellow!15}4.09 & \cellcolor{yellow!15}8.23 & \cellcolor{yellow!15}9.50 & \cellcolor{orange!15}\textbf{7.27} & \cellcolor{yellow!15}81.50 & \cellcolor{yellow!15}27.63 & \cellcolor{orange!15}\textbf{54.57} & \cellcolor{yellow!15}56.10 & \cellcolor{yellow!15}63.80 & \cellcolor{orange!15}\textbf{59.95} & \cellcolor{yellow!15}80.26 & \cellcolor{yellow!15}60.32 & \cellcolor{yellow!15}85.02 & \cellcolor{yellow!15}48.93 & \cellcolor{yellow!15}65.39 & \cellcolor{yellow!15}82.70 & \cellcolor{yellow!15}79.32 & \cellcolor{orange!15}\textbf{71.71} \\ \midrule
 \multirow{11}{*}{\textbf{Qwen3-32B}} & \cellcolor{gray!15}\textbf{0\%} & \cellcolor{gray!15}\textbf{N/A} & \cellcolor{gray!15}7.19 & \cellcolor{gray!15}13.51 & \cellcolor{gray!15}11.97 & \cellcolor{orange!15}10.89 & \cellcolor{gray!15}79.40 & \cellcolor{gray!15}33.48 & \cellcolor{orange!15}56.44 & \cellcolor{gray!15}79.88 & \cellcolor{gray!15}64.80 & \cellcolor{orange!15}72.34 & \cellcolor{gray!15}81.88 & \cellcolor{gray!15}58.45 & \cellcolor{gray!15}84.13 & \cellcolor{gray!15}52.41 & \cellcolor{gray!15}63.86 & \cellcolor{gray!15}81.28 & \cellcolor{gray!15}73.16 & \cellcolor{orange!15}70.74 \\ \cmidrule{2-21}
 & \multirow{5}{*}{\textbf{\shortstack{25\%\\Layer}}} & \textbf{ShortGPT} & 12.72 & 25.37 & 22.28 & \cellcolor{orange!15}20.12 & 20.70 & 6.76 & \cellcolor{orange!15}13.73 & 8.54 & 15.60 & \cellcolor{orange!15}12.07 & 44.22 & 38.57 & 70.54 & 41.20 & 45.56 & 72.47 & 53.99 & \cellcolor{orange!15}52.36 \\
 &  & \textbf{LaCo} & 10.46 & 18.76 & 17.47 & \cellcolor{orange!15}15.56 & 57.62 & 24.50 & \cellcolor{orange!15}41.06 & 53.05 & 11.80 & \cellcolor{orange!15}32.43 & 65.18 & 49.74 & 79.92 & 45.80 & 55.26 & 74.70 & 65.35 & \cellcolor{orange!15}62.28 \\
 &  & \textbf{LLM-Pruner} & 7.68 & 13.02 & 13.05 & \cellcolor{orange!15}\textbf{11.25} & 37.38 & 19.50 & \cellcolor{orange!15}28.44 & 6.10 & 19.20 & \cellcolor{orange!15}12.65 & 76.64 & 22.70 & 25.08 & 47.29 & 60.29 & 49.51 & 69.69 & \cellcolor{orange!15}50.17 \\
 &  & \textbf{EvoPress} & 9.70 & 16.76 & 15.89 & \cellcolor{orange!15}14.12 & 73.69 & 22.30 & \cellcolor{orange!15}48.00 & 39.02 & 54.80 & \cellcolor{orange!15}46.91 & 69.08 & 52.73 & 79.71 & 46.11 & 53.85 & 75.46 & 65.11 & \cellcolor{orange!15}63.15 \\
 &  & \cellcolor{yellow!15}\textbf{SliMOO} & \cellcolor{yellow!15}9.62 & \cellcolor{yellow!15}16.62 & \cellcolor{yellow!15}15.62 & \cellcolor{orange!15}{\ul 13.95} & \cellcolor{yellow!15}75.89 & \cellcolor{yellow!15}24.40 & \cellcolor{orange!15}\textbf{50.15} & \cellcolor{yellow!15}57.32 & \cellcolor{yellow!15}56.80 & \cellcolor{orange!15}\textbf{57.06} & \cellcolor{yellow!15}68.21 & \cellcolor{yellow!15}53.58 & \cellcolor{yellow!15}79.29 & \cellcolor{yellow!15}46.57 & \cellcolor{yellow!15}53.71 & \cellcolor{yellow!15}75.35 & \cellcolor{yellow!15}66.30 & \cellcolor{orange!15}\textbf{63.29} \\ \cmidrule{2-21}
 & \multirow{5}{*}{\textbf{\shortstack{33\%\\Sublayer}}} & \textbf{ShortGPT} & 13.51 & 26.10 & 24.28 & \cellcolor{orange!15}21.30 & 45.79 & 16.72 & \cellcolor{orange!15}31.26 & 42.68 & 27.80 & \cellcolor{orange!15}35.24 & 44.88 & 42.41 & 70.50 & 42.22 & 45.66 & 69.70 & 56.91 & \cellcolor{orange!15}53.18 \\
 &  & \textbf{EvoPress} & 9.81 & 18.62 & 16.88 & \cellcolor{orange!15}\textbf{15.10} & 56.10 & 15.06 & \cellcolor{orange!15}35.58 & 18.90 & 37.60 & \cellcolor{orange!15}28.25 & 57.79 & 42.32 & 71.84 & 41.97 & 50.79 & 73.12 & 58.64 & \cellcolor{orange!15}56.64 \\
 &  & \textbf{BlockPruner} & 12.32 & 23.17 & 19.81 & \cellcolor{orange!15}18.43 & 44.43 & 14.34 & \cellcolor{orange!15}29.39 & 19.51 & 30.20 & \cellcolor{orange!15}24.86 & 50.56 & 43.09 & 73.78 & 42.78 & 48.63 & 72.85 & 60.85 & \cellcolor{orange!15}56.08 \\
 &  & \textbf{FinerCut} & 19.49 & 43.06 & 26.76 & \cellcolor{orange!15}29.77 & 1.30 & 1.04 & \cellcolor{orange!15}1.17 & 3.05 & 4.00 & \cellcolor{orange!15}3.53 & 46.02 & 33.87 & 57.41 & 42.53 & 46.99 & 71.16 & 55.64 & \cellcolor{orange!15}50.52 \\
 &  & \cellcolor{yellow!15}\textbf{SliMOO} & \cellcolor{yellow!15}10.75 & \cellcolor{yellow!15}20.75 & \cellcolor{yellow!15}18.00 & \cellcolor{orange!15}{\ul 16.50} & \cellcolor{yellow!15}58.30 & \cellcolor{yellow!15}20.64 & \cellcolor{orange!15}\textbf{39.47} & \cellcolor{yellow!15}46.34 & \cellcolor{yellow!15}46.00 & \cellcolor{orange!15}\textbf{46.17} & \cellcolor{yellow!15}65.90 & \cellcolor{yellow!15}47.27 & \cellcolor{yellow!15}76.77 & \cellcolor{yellow!15}44.42 & \cellcolor{yellow!15}51.19 & \cellcolor{yellow!15}75.14 & \cellcolor{yellow!15}62.67 & \cellcolor{orange!15}\textbf{60.48} \\ \midrule
\multirow{10}{*}{\textbf{Qwen3-235B-A22B}} & \cellcolor{gray!15}\textbf{0\%} & \cellcolor{gray!15}\textbf{N/A} & \cellcolor{gray!15}5.63 & \cellcolor{gray!15}11.58 & \cellcolor{gray!15}10.39 & \cellcolor{orange!15}9.20 & \cellcolor{gray!15}84.99 & \cellcolor{gray!15}42.48 & \cellcolor{orange!15}63.74 & \cellcolor{gray!15}79.27 & \cellcolor{gray!15}79.60 & \cellcolor{orange!15}79.44 & \cellcolor{gray!15}86.35 & \cellcolor{gray!15}61.86 & \cellcolor{gray!15}85.27 & \cellcolor{gray!15}54.96 & \cellcolor{gray!15}67.88 & \cellcolor{gray!15}81.99 & \cellcolor{gray!15}78.93 & \cellcolor{orange!15}73.89 \\ \cmidrule{2-21}
 & \multirow{4}{*}{\textbf{\shortstack{25\%\\Layer}}} & \textbf{ShortGPT} & 6.07 & 12.16 & 11.83 & \cellcolor{orange!15}10.02 & 84.23 & 35.82 & \cellcolor{orange!15}60.03 & 31.71 & 74.40 & \cellcolor{orange!15}53.06 & 79.33 & 56.91 & 83.04 & 50.15 & 62.78 & 78.94 & 73.80 & \cellcolor{orange!15}69.28 \\
 &  & \textbf{LaCo} & 6.22 & 13.37 & 11.60 & \cellcolor{orange!15}10.40 & 81.80 & 36.56 & \cellcolor{orange!15}59.18 & 31.10 & 71.80 & \cellcolor{orange!15}51.45 & 79.69 & 57.68 & 83.21 & 50.87 & 63.40 & 79.22 & 74.90 & \cellcolor{orange!15}69.85 \\
 &  & \textbf{EvoPress} & 5.72 & 11.44 & 10.94 & \cellcolor{orange!15}\textbf{9.37} & 82.64 & 38.26 & \cellcolor{orange!15}60.45 & 34.76 & 74.80 & \cellcolor{orange!15}54.78 & 82.38 & 58.28 & 82.53 & 52.81 & 64.93 & 80.69 & 75.77 & \cellcolor{orange!15}71.06 \\
 &  & \cellcolor{yellow!15}\textbf{SliMOO} & \cellcolor{yellow!15}5.72 & \cellcolor{yellow!15}11.44 & \cellcolor{yellow!15}11.06 & \cellcolor{orange!15}{\ul 9.41} & \cellcolor{yellow!15}84.38 & \cellcolor{yellow!15}40.00 & \cellcolor{orange!15}\textbf{62.19} & \cellcolor{yellow!15}35.98 & \cellcolor{yellow!15}76.80 & \cellcolor{orange!15}\textbf{56.39} & \cellcolor{yellow!15}85.53 & \cellcolor{yellow!15}62.97 & \cellcolor{yellow!15}85.44 & \cellcolor{yellow!15}52.87 & \cellcolor{yellow!15}65.12 & \cellcolor{yellow!15}80.41 & \cellcolor{yellow!15}76.80 & \cellcolor{orange!15}\textbf{72.73} \\ \cmidrule{2-21}
 & \multirow{5}{*}{\textbf{\shortstack{33\%\\Sublayer}}} & \textbf{ShortGPT} & 6.18 & 12.35 & 12.23 & \cellcolor{orange!15}10.25 & 80.82 & 36.30 & \cellcolor{orange!15}58.56 & 35.98 & 71.80 & \cellcolor{orange!15}53.89 & 77.47 & 53.67 & 81.10 & 49.54 & 61.74 & 80.09 & 73.88 & \cellcolor{orange!15}68.21 \\
 &  & \textbf{EvoPress} & 5.93 & 11.67 & 11.02 & \cellcolor{orange!15}\textbf{9.54} & 72.86 & 25.18 & \cellcolor{orange!15}49.02 & 26.22 & 52.20 & \cellcolor{orange!15}39.21 & 75.44 & 54.44 & 80.93 & 51.79 & 59.92 & 79.33 & 72.61 & \cellcolor{orange!15}67.78 \\
 &  & \textbf{BlockPruner} & 7.37 & 12.95 & 12.67 & \cellcolor{orange!15}11.00 & 60.05 & 25.10 & \cellcolor{orange!15}42.58 & 37.80 & 51.20 & \cellcolor{orange!15}44.50 & 73.46 & 53.67 & 81.10 & 50.05 & 59.63 & 78.56 & 71.35 & \cellcolor{orange!15}66.83 \\
 &  & \textbf{FinerCut} & 6.75 & 14.98 & 12.40 & \cellcolor{orange!15}11.38 & 65.35 & 26.38 & \cellcolor{orange!15}45.87 & 31.71 & 43.40 & \cellcolor{orange!15}37.56 & 74.06 & 54.27 & 79.00 & 48.87 & 60.47 & 78.78 & 71.43 & \cellcolor{orange!15}66.70 \\
 &  & \cellcolor{yellow!15}\textbf{SliMOO} & \cellcolor{yellow!15}5.94 & \cellcolor{yellow!15}11.81 & \cellcolor{yellow!15}11.81 & \cellcolor{orange!15}{\ul 9.85} & \cellcolor{yellow!15}80.67 & \cellcolor{yellow!15}38.70 & \cellcolor{orange!15}\textbf{59.69} & \cellcolor{yellow!15}39.63 & \cellcolor{yellow!15}74.40 & \cellcolor{orange!15}\textbf{57.02} & \cellcolor{yellow!15}82.57 & \cellcolor{yellow!15}59.64 & \cellcolor{yellow!15}83.84 & \cellcolor{yellow!15}50.26 & \cellcolor{yellow!15}63.08 & \cellcolor{yellow!15}80.03 & \cellcolor{yellow!15}74.35 & \cellcolor{orange!15}\textbf{70.54} \\ \bottomrule
\end{tabular}}
\end{sidewaystable}

The conservative prior $\hat I_i=\max_t I_{t,i}$ follows from the multi-objective nature of the problem. A unit with a large score for any task is potentially capability-critical, even if it appears unimportant for other tasks. Using the maximum therefore protects task-specific layers that would be washed out by averaging. At the same time, SliMOO also samples task-specific rows $I_{t,:}$ during mutation, because a purely conservative prior would over-protect shared high-sensitivity regions and could under-explore domain-specific Pareto trade-offs. In this sense, the balanced prior controls risk, while task-specific priors maintain search diversity.

Feasibility is preserved by viewing masks in $\mathcal{X}_K$ as $K$-subsets of the feasible unit set $\mathcal F$. The most local valid move is a swap: remove one currently kept unit from $\mathcal F$ and restore one currently pruned unit. This changes the mask by Hamming distance two but keeps $\|\mathbf{x}\|_1=K$, so every mutated candidate remains at the target sparsity. TLIP enters this move only through the proposal distribution: low-prior kept units are more likely to be newly pruned, and high-prior removed units are more likely to be restored. Cardinality-preserving crossover serves the same purpose at the population level. Given two parent pruning sets $S_1$ and $S_2$ with $|S_1|=|S_2|=K$, the offspring first inherits their shared pruned units $S_1\cap S_2$, and then fills the remaining slots by sampling from the symmetric difference $(S_1\cup S_2)\setminus(S_1\cap S_2)$ until its size reaches $K$. This recombines useful pruning patterns from both parents while guaranteeing that the child remains in $\mathcal{X}_K$.

After candidate generation, selection is performed in objective space. Let $\mathcal C$ denote the candidate pool formed by the current population and newly generated offspring. NSGA-II constructs non-dominated fronts recursively:
\begin{equation}
\mathcal F_1=\{\mathbf{x}\in\mathcal C:\nexists\,\mathbf{y}\in\mathcal C,\ \mathbf{y}\prec\mathbf{x}\},
\qquad
\mathcal F_{\ell}=\mathrm{ND}\!\left(\mathcal C\setminus\bigcup_{j<\ell}\mathcal F_j\right),\ \ell\ge 2,
\end{equation}
where $\mathbf{x}$ and $\mathbf{y}$ are candidate masks, $\prec$ denotes Pareto dominance, $\mathcal F_\ell$ is the $\ell$-th front, and $\mathrm{ND}(\cdot)$ returns the non-dominated subset of its input set. Front rank gives a coarse ordering of solution quality: all masks in $\mathcal F_1$ are Pareto-competitive within the pool, while later fronts are dominated by earlier ones. When the population budget cuts through a front, NSGA-II sorts candidates in that front by crowding distance in descending order and retains those with larger distances first, with boundary solutions treated as maximally distant. This rule favors solutions located in sparse regions of objective space, preserving coverage of different trade-offs and preventing the population from collapsing to a narrow set of masks that favor only one capability domain.

After search, the final population contains multiple non-dominated masks with different trade-offs, but deployment requires selecting a single pruning decision. SliMOO therefore uses hypervolume contribution to choose the final mask. Let $\mathcal P$ denote the final Pareto population and set the reference point by $r_t=1.1\max_{\mathbf{x}\in\mathcal A} f_t(\mathbf{x})$, where $\mathcal A$ is the highest-fidelity evaluated archive. For minimization objectives, the hypervolume is the dominated volume:
\begin{equation}
\mathrm{HV}(\mathcal P;\mathbf r)
=
\lambda\!\left(
\bigcup_{\mathbf{x}\in\mathcal P}
[\mathbf f(\mathbf{x}),\mathbf r]
\right),
\end{equation}
where $\lambda(\cdot)$ denotes volume and $[\mathbf f(\mathbf{x}),\mathbf r]$ is the axis-aligned box from candidate $\mathbf{x}$ to the reference point. The marginal contribution of a candidate $\mathbf{x}\in\mathcal P$ is:
\begin{equation}
\Delta_{\mathrm{HV}}(\mathbf{x};\mathcal P)
=
\mathrm{HV}(\mathcal P;\mathbf r)-\mathrm{HV}(\mathcal P\setminus\{\mathbf{x}\};\mathbf r).
\end{equation}
A large contribution indicates that the candidate expands the covered Pareto region rather than merely duplicating nearby trade-offs. Thus, hypervolume provides a principled tie-breaker for selecting a compact final mask from a diverse Pareto population. Overall, the derivation shows how SliMOO connects three levels of reasoning: TLIP provides a first-order task-layer prior, NSGA-II converts task-wise losses into diversity-preserving Pareto selection, and hypervolume identifies a final solution with strong global contribution to the trade-off surface.

\textbf{Multi-stage evaluation for efficient search}. To avoid evaluating every candidate with the full calibration budget, we use a successive schedule inside the objective evaluation step of Algorithm~\ref{alg:SliMOO}. Candidate pools are filtered by NSGA-II with token budgets $2{,}048\rightarrow4{,}096\rightarrow16{,}384$ and survivor counts $16\rightarrow8\rightarrow 4$; the same schedule is used for initialization. This accelerates search by discarding weak masks at low fidelity while leaving candidate generation and Pareto selection unchanged.

\subsection{Detailed Post-pruning Performance}
\label{detailed}

Table~\ref{tab:detailed-post-pruning} expands the domain averages in the main text into a benchmark-level view, making it easier to distinguish broad capability preservation from gains concentrated on a single metric. Across all models, SliMOO is consistently strongest on the math and code averages while remaining competitive on language modeling and knowledge reasoning. This pattern is similarly clear on Qwen family models, where the benchmark-level results show gains across multiple math and code benchmarks rather than a single outlier score.

Consistent with this observation, on Qwen3-32B at $25\%$ sparsity, relative to EvoPress, SliMOO improves GSM8K/MATH from $73.69\%/22.30\%$ to $75.89\%/24.40\%$ and HumanEval/MBPP from $39.02\%/54.80\%$ to $57.32\%/56.80\%$, while slightly improving the language modeling average from $14.12$ to $13.95$. The same trend is even stronger on Qwen3-235B-A22B at $33\%$ sparsity, where SliMOO improves GSM8K/MATH from $72.86\%/25.18\%$ to $80.67\%/38.70\%$ and HumanEval/MBPP from $26.22\%/52.20\%$ to $39.63\%/74.40\%$, while also raising the knowledge reasoning average from $67.78\%$ to $70.54\%$; the Llama experiments show the same direction with gains across all domains. Together, these results support the paper's central claim that task-aware, Pareto-guided pruning preserves broad capability more reliably than a single global pruning order.

\begin{sidewaystable}[p]
\centering
\caption{Layer-wise pruning configurations selected by SliMOO.
Each column denotes one transformer layer, with attention on the
top strip and MLP on the bottom strip. Salmon blocks are retained
and white blocks are pruned; arrows point from shallow to deep layers.}
\label{tab:pruning-mask}

\includegraphics[width=\textheight]{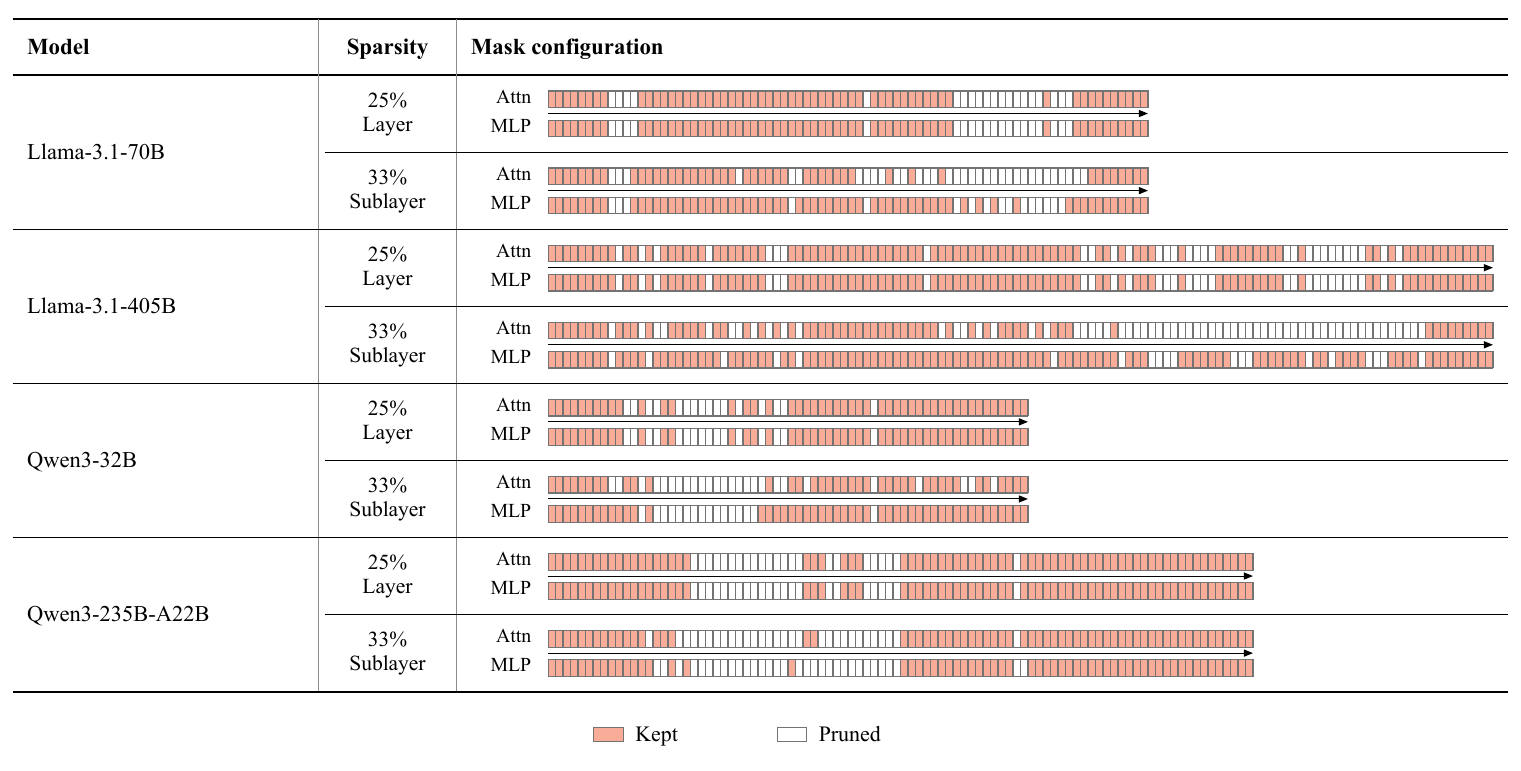}

\end{sidewaystable}

\subsection{Pruning and Tuning}
\label{tune}

\begin{wraptable}{r}{0.5\textwidth}
\vspace{-1.7em}
\centering
\caption{Post-pruning LoRA tuning results on Llama-3.1-70B at $25\%$ layer sparsity. Starred methods are tuned after pruning with LoRA.}
\label{tab:post-pruning-lora}
\setlength{\tabcolsep}{4pt}
\renewcommand{\arraystretch}{1.08}
\footnotesize
\resizebox{\linewidth}{!}{
\begin{tabular}{l|cccc}
\toprule
Method & Lang. $\downarrow$ & Math $\uparrow$ & Code $\uparrow$ & Know. $\uparrow$ \\
\midrule
ShortGPT & 11.35 & 12.85 & 15.45 & 65.93 \\
ShortGPT$^\ast$ & 9.13 & 37.35 & 49.18 & 68.31 \\ \midrule
EvoPress & 9.88 & 24.33 & 35.03 & 64.63 \\
EvoPress$^\ast$ & \textbf{8.48} & 42.22 & 57.23 & 68.32 \\ \midrule
\cellcolor{yellow!15}SliMOO & \cellcolor{yellow!15}9.89 & \cellcolor{yellow!15}34.40 & \cellcolor{yellow!15}39.14 & \cellcolor{yellow!15}66.90 \\
\cellcolor{orange!15}SliMOO$^\ast$ & \cellcolor{orange!15}8.56 & \cellcolor{orange!15}\textbf{48.04} & \cellcolor{orange!15}\textbf{61.40} & \cellcolor{orange!15}\textbf{70.23} \\
\bottomrule
\end{tabular}}
\vspace{-1.0em}
\end{wraptable}

Post-pruning fine-tuning is commonly used to recover compression-induced capability loss with controllable additional cost. Table~\ref{tab:post-pruning-lora} evaluates this setting with lightweight LoRA tuning~\cite{lora} after depth pruning. Starting from the pruned Llama-3.1-70B checkpoints, we tune all linear layers with LoRA rank $16$, $\alpha=32$, dropout $0.05$, learning rate $1.0\times10^{-4}$, cosine scheduling, and warmup ratio $0.03$. Training uses roughly $25\%$ of \texttt{tulu-3-sft-mixture}, a broad supervised instruction-tuning mixture containing about $939$K examples from diverse sources, including general instruction following, math, code, multilingual, and domain-specific data~\cite{tulu}. Finetuning improves all pruned models, but SliMOO remains the strongest: SliMOO$^\ast$ achieves the best math, code, and knowledge averages while matching the best language-modeling recovery closely. This suggests that SliMOO's pruning mask preserves a more tunable residual model than masks selected by single-score or scalarized search baselines.

\subsection{Small Model Experiments}
\label{sec:small}

\begin{wraptable}{r}{0.58\textwidth}
\vspace{-1.7em}
\centering
\caption{Small-model pruning results on Llama-3.1-8B. We report domain averages: lower is better for language modeling and higher is better elsewhere. \textbf{Bold} marks the best pruned model within each sparsity setting.}
\label{tab:small-model-results}
\setlength{\tabcolsep}{3.5pt}
\renewcommand{\arraystretch}{1.06}
\footnotesize
\resizebox{\linewidth}{!}{
\begin{tabular}{c|l|cccc}
\toprule
Sparsity & Method & Lang. $\downarrow$ & Math $\uparrow$ & Code $\uparrow$ & Know. $\uparrow$ \\
\midrule
\cellcolor{gray!15}$0\%$ & \cellcolor{gray!15}N/A & \cellcolor{gray!15}8.47 & \cellcolor{gray!15}32.10 & \cellcolor{gray!15}42.20 & \cellcolor{gray!15}66.04 \\
\midrule
\multirow{4}{*}{\shortstack{$12.5\%$\\Layer}}
& ShortGPT & 17.60 & 19.17 & 8.37 & 61.28 \\
& LaCo & 12.73 & 21.02 & 17.93 & 60.13 \\
& EvoPress & 19.54 & 1.27 & 3.02 & 48.40 \\
& \cellcolor{yellow!15}SliMOO & \cellcolor{yellow!15}\textbf{11.65} & \cellcolor{yellow!15}\textbf{23.30} & \cellcolor{yellow!15}\textbf{23.26} & \cellcolor{yellow!15}\textbf{61.64} \\
\midrule
\multirow{5}{*}{\shortstack{$25\%$\\Sublayer}}
& ShortGPT & 280.00 & 0.02 & 0.00 & 46.89 \\
& EvoPress & 24.48 & 1.68 & 2.17 & 47.32 \\
& BlockPruner & 15.94 & 2.80 & 3.65 & 54.63 \\
& FinerCut & \textbf{15.12} & 5.30 & 5.74 & 53.73 \\
& \cellcolor{yellow!15}SliMOO & \cellcolor{yellow!15}15.58 & \cellcolor{yellow!15}\textbf{8.46} & \cellcolor{yellow!15}\textbf{9.47} & \cellcolor{yellow!15}\textbf{55.04} \\
\bottomrule
\end{tabular}}
\vspace{-1.0em}
\end{wraptable}

As model scale increases, depth pruning can exploit parameter redundancy that is less pronounced in smaller models. We therefore additionally evaluate Llama-3.1-8B as a small-model setting, where pruning is expected to be more fragile and use milder budgets: $12.5\%$ whole-layer pruning and $25\%$ attention/MLP sublayer pruning. Table~\ref{tab:small-model-results} shows that SliMOO remains effective in this more sensitive regime. Under $12.5\%$ layer sparsity, SliMOO achieves the best pruned result on all four domain averages, indicating that task-aware search can identify removable layers even when redundancy is limited. Under $25\%$ sublayer pruning, SliMOO obtains the best math, code, and knowledge averages and remains competitive on language modeling, while several baselines suffer severe capability collapse. These results suggest that SliMOO's multi-objective formulation is useful not only for highly redundant large models but also for smaller models where pruning decisions must be more conservative.

\subsection{Pruning Position and Interpretability}
\label{interpretability}

Table~\ref{tab:pruning-mask} visualizes the final pruning decisions selected by SliMOO across model families and pruning granularities. TLIP provides a task-aware prior that biases search away from task-critical units, while Pareto evaluation determines which combinations are actually retained. This makes the final masks interpretable as search-refined versions of the task-layer importance patterns in Figure~\ref{fig:slimoo}(a). For example, under $33\%$ sublayer pruning on Llama-3.1-70B, SliMOO keeps most early and final sublayers, but prunes a concentrated mid-to-late region, especially attention sublayers from roughly layers $41$--$71$, with selected MLP removals interleaved around layers $42$ and $54$--$68$. This pattern is consistent with TLIP: broadly important units are protected, while lower-sensitivity or task-redundant regions provide pruning capacity. At the same time, the mask is not purely local or monotonic, reflecting the role of Pareto search in accounting for cross-task interactions among removed sublayers.

\subsection{Robustness to Search Random Seeds}
\label{error-bar}

\begin{wraptable}{r}{0.55\textwidth}
\vspace{-1em}
\centering
\footnotesize
\setlength{\tabcolsep}{3pt}
\renewcommand{\arraystretch}{1.15}
\caption{Search-seed stability on Llama-3.1-70B at $33\%$ sublayer sparsity. The last column reports mean $\pm$ SD across five runs. Language reports perplexity; other scores are percentages.}
\label{tab:multi-seed}
\begin{tabular*}{\linewidth}{@{\extracolsep{\fill}}lccc@{}}
\toprule
Domain
& EvoPress
& \shortstack{SliMOO reported}
& \shortstack{SliMOO 5 runs} \\
\midrule
Language $\downarrow$
& 12.02 & 10.45 & $10.34 \pm 0.14$ \\
Math $\uparrow$
& 10.33 & 27.27 & $27.31 \pm 0.17$ \\
Code $\uparrow$
& 15.11 & 30.78 & $30.56 \pm 0.45$ \\
Knowledge $\uparrow$
& 62.02 & 66.29 & $66.19 \pm 0.07$ \\
\bottomrule
\end{tabular*}
\end{wraptable}

We assess the sensitivity of SliMOO to search randomness on Llama-3.1-70B. At $25\%$ layer sparsity, all three runs converge to the same pruning mask and yield identical domain averages: $9.89$ perplexity for language modeling and $34.40\%$, $39.14\%$, and $66.90\%$ for math, code, and knowledge reasoning, respectively.
At $33\%$ sublayer sparsity, we evaluate five search seeds, including the reported run. As shown in Table~\ref{tab:multi-seed}, the domain averages exhibit limited variation across runs. These results support the robustness of SliMOO to search randomness at both pruning granularities in the evaluated setting.

\subsection{Case Study}
\label{case}

\begin{table}[t]
\centering
\caption{Case study on Llama-3.1-70B before and after dropping 25\% of layers with SliMOO. We use greedy decoding for generation. For math, code, and reasoning, decoding is truncated after the final answer.}
\label{tab:case-study}
\setlength{\tabcolsep}{5pt}
\renewcommand{\arraystretch}{1.18}
\footnotesize
\resizebox{\linewidth}{!}{
\begin{tabular}{p{0.11\textwidth}|p{0.23\textwidth}|p{0.30\textwidth}|p{0.30\textwidth}}
\toprule
\textbf{Domain} & \textbf{Prompt (truncated)} & \textbf{Dense output} & \textbf{Pruned output} \\
\midrule
\textbf{Language modeling} &
\textit{``The Great Wall of China is a series of fortifications that were built across the historical northern borders of ancient Chinese states and Imperial China as protection against various nomadic groups. Construction of the walls began as early as the 7th century BC, and''} &
they have a total length of 13{,}000 miles. Several sections of the wall are now in ruins or have disappeared. However, the wall is still one of the most attractive places in the world and was selected as one of the New Seven Wonders of the World in 2007. &
the most recent construction was completed in the 18th century. The walls were constructed of stone, compacted earth, and compacted sand, depending on the area. The walls were constructed on high ground, and they overlook nearby plains. The walls were constructed with the idea of protecting the states from raiders, and they were constructed with the assumption that they could be defended by a minimal force. \\
\midrule
\textbf{Math} &
\textit{``Janet's ducks lay 16 eggs per day. She eats three for breakfast every morning and bakes muffins for her friends every day with four. She sells the remainder at the farmers' market daily for \$2 per fresh duck egg. How much in dollars does she make every day at the farmers' market?''} &
Janet's ducks lay 16 eggs per day. She eats three for breakfast every morning and bakes muffins for her friends every day with four. She sells the remainder at the farmers' market daily for \$2 per fresh duck egg. $16-3-4=9$. $9\times 2=18$. \textbf{The answer is \$18.} $\checkmark$ &
Janet's ducks lay $16-3-4=9$ eggs per day. She sells 9 eggs per day. She makes $9\times 2=\$18$ per day. \textbf{The answer is \$18.} $\checkmark$ \\
\midrule
\textbf{Code} &
\texttt{def is\_prime(n: int) -> bool:}\newline
\texttt{"""Return True if n is prime."""} &
\begin{tabular}[t]{@{}l@{}}
\texttt{if n < 2:}\\
\quad \texttt{return False}\\
\texttt{for i in range(2,}\texttt{\textbackslash}\\
\hspace*{4em}\texttt{int(n ** 0.5) + 1):}\\
\quad \texttt{if n \% i == 0:}\\
\qquad \texttt{return False}\\
\texttt{return True}
\end{tabular}\newline
\textbf{Optimal $O(\sqrt{n})$.} $\checkmark$ &
\begin{tabular}[t]{@{}l@{}}
\texttt{if n < 2:}\\
\quad \texttt{return False}\\
\texttt{for i in range(2, n):}\\
\quad \texttt{if n \% i == 0:}\\
\qquad \texttt{return False}\\
\texttt{return True}
\end{tabular}\newline
Functionally correct but \textbf{naive $O(n)$}; the $\sqrt{n}$ optimization is lost. $\checkmark$ \\
\midrule
\textbf{Knowledge reasoning} &
\parbox[t]{\linewidth}{
\textit{``In which year did World War II end?}\\
\textit{A.~1943 \quad B.~1944}\\
\textit{C.~1945 \quad D.~1946}\\
\textit{Answer:''}
} &
\textbf{C} $\checkmark$ &
\textbf{C} $\checkmark$ \\
\bottomrule
\end{tabular}}
\end{table}

Table~\ref{tab:case-study} provides a qualitative comparison between the dense and SliMOO-pruned Llama-3.1-70B on representative prompts from the four capability domains. Overall, both models produce strong outputs, indicating that the pruned model retains the core capabilities of the dense model after removing $25\%$ of the layers. The differences are mostly subtle: in language modeling, the pruned continuation remains fluent and coherent; in math and knowledge reasoning, it preserves the correct final answer; and in code, it remains functionally correct although less algorithmically refined than the dense model. These examples qualitatively reinforce the main finding of the paper that SliMOO preserves broad capability across diverse domains after depth pruning.

\subsection{Existing Assets and Licenses}
\label{assets-licenses}

Table~\ref{tab:assets-licenses} lists the main existing assets used in our experiments and the licenses reported by their source pages.

\begin{table}[h!]
\centering
\caption{Existing models, datasets, and benchmarks used in the paper.}
\label{tab:assets-licenses}
\footnotesize
\setlength{\tabcolsep}{3pt}
\renewcommand{\arraystretch}{1.05}
\resizebox{\linewidth}{!}{
\begin{tabular}{p{0.24\linewidth}p{0.24\linewidth}p{0.47\linewidth}}
\toprule
Asset & License & Source \\
\midrule
Llama-3.1~\cite{llama3} & Llama 3.1 Community & \url{https://llama.com/llama3_1/license/} \\
Qwen3~\cite{qwen3} & Apache-2.0 & \url{https://huggingface.co/Qwen/Qwen3-32B/blob/main/LICENSE}; \url{https://huggingface.co/Qwen/Qwen3-235B-A22B/blob/main/LICENSE} \\
FineWeb-Edu~\cite{fineweb-edu} & ODC-BY-1.0 & \url{https://huggingface.co/datasets/HuggingFaceFW/fineweb-edu} \\
FineWeb-2~\cite{fineweb2} & ODC-BY-1.0 & \url{https://huggingface.co/datasets/HuggingFaceFW/fineweb-2} \\
GSM8K~\cite{gsm8k} & MIT & \url{https://huggingface.co/datasets/openai/gsm8k} \\
DeepScaleR~\cite{deepscaler} & MIT & \url{https://huggingface.co/datasets/agentica-org/DeepScaleR-Preview-Dataset} \\
python-codes-25k~\cite{pythoncodes25k} & MIT & \url{https://huggingface.co/datasets/flytech/python-codes-25k} \\
commonsense\_170k~\cite{dora} & Apache-2.0 & \url{https://huggingface.co/datasets/zwhe99/commonsense_170k}; \url{https://github.com/AGI-Edgerunners/LLM-Adapters} \\
Tulu-3 SFT mixture~\cite{tulu} & ODC-BY-1.0 & \url{https://huggingface.co/datasets/allenai/tulu-3-sft-mixture} \\
WikiText-2~\cite{wikitext2} & CC-BY-SA-3.0 / GFDL & \url{https://huggingface.co/datasets/Salesforce/wikitext} \\
PTB~\cite{ptb} & LDC User Agreement & \url{https://catalog.ldc.upenn.edu/LDC99T42} \\
C4~\cite{c4} & ODC-BY-1.0 & \url{https://huggingface.co/datasets/allenai/c4} \\
MATH~\cite{math} & MIT & \url{https://github.com/hendrycks/math} \\
HumanEval~\cite{humaneval} & MIT & \url{https://github.com/openai/human-eval} \\
MBPP~\cite{mbpp} & CC-BY-4.0 & \url{https://huggingface.co/datasets/google-research-datasets/mbpp} \\
MMLU~\cite{mmlu} & MIT & \url{https://huggingface.co/datasets/cais/mmlu} \\
ARC-C/E~\cite{arc} & CC-BY-SA-4.0 & \url{https://huggingface.co/datasets/allenai/ai2_arc} \\
Social-IQA~\cite{siqa} & CC-BY-4.0 & \url{https://huggingface.co/datasets/allenai/social_i_qa} \\
HellaSwag~\cite{hellaswag} & MIT & \url{https://github.com/rowanz/hellaswag} \\
PIQA~\cite{piqa} & CC-BY-SA-4.0 & \url{https://huggingface.co/datasets/ybisk/piqa} \\
WinoGrande~\cite{winogrande} & Apache-2.0 & \url{https://github.com/allenai/winogrande} \\
RULER (NIAH-8K)~\cite{ruler} & Apache-2.0 & \url{https://github.com/NVIDIA/RULER} \\
TruthfulQA~\cite{truthfulqa} & Apache-2.0 & \url{https://huggingface.co/datasets/truthfulqa/truthful_qa} \\
CrowS-Pairs~\cite{crows-pairs} & CC-BY-SA-4.0 & \url{https://github.com/nyu-mll/crows-pairs} \\
ToxiGen~\cite{toxigen} & CDLA-Permissive-2.0 & \url{https://huggingface.co/datasets/toxigen/toxigen-data}; \url{https://github.com/microsoft/TOXIGEN/blob/main/LICENSE.txt} \\
MedMCQA~\cite{medmcqa} & Apache-2.0 & \url{https://huggingface.co/datasets/openlifescienceai/medmcqa} \\
\bottomrule
\end{tabular}}
\end{table}

\subsection{Broader Impacts}\label{broader-impact}

SliMOO has primarily positive broader impacts by making large language models more efficient to deploy. By reducing model depth while preserving broad capabilities, it can lower inference latency, memory footprint, and serving cost, making capable LLMs more accessible under practical resource constraints. These efficiency gains may also reduce the energy and hardware requirements of deployment. SliMOO does not introduce new data, new model capabilities, or new high-risk modalities beyond the scope of the pre-trained models it compresses. However, pruning may affect safety guardrails or instruction alignment unevenly, so deployment should evaluate relevant safety and alignment benchmarks when these properties are required.


\end{document}